\documentclass{article} % For LaTeX2e
\usepackage[submission]{colm2025_conference}
\usepackage{microtype}
\usepackage{hyperref}
\usepackage{url}
\usepackage{booktabs}
\usepackage{comment}
\usepackage{amsmath}
\usepackage{algorithm}
\usepackage[noend]{algpseudocode}
\usepackage{subcaption}
\usepackage{graphicx}
\usepackage{lineno}
\usepackage{makecell} % for multi-line cells
\usepackage{tabularx} % for fixed-width columns with wrapping
\usepackage{booktabs} % for better table lines
\usepackage{multirow}

\definecolor{darkblue}{rgb}{0, 0, 0.5}
\hypersetup{colorlinks=true, citecolor=darkblue, linkcolor=darkblue, urlcolor=darkblue}

\usepackage{listings}
\usepackage{xcolor}
\usepackage{wrapfig}  % Add this to your preamble if it's not already included

\definecolor{promptbg}{RGB}{240,240,240}
\definecolor{promptframe}{RGB}{204,204,204}

\lstdefinestyle{promptstyle}{
    backgroundcolor=\color{promptbg},
    frame=single,
    framesep=3pt,
    rulecolor=\color{promptframe},
    basicstyle=\ttfamily\small,
    breaklines=true,
    breakatwhitespace=true,
    showstringspaces=false,
    columns=flexible,
    keepspaces=true,
    numbers=none,
    commentstyle=\color{gray},
    keywordstyle=\color{black},
    stringstyle=\color{black},
    captionpos=b
}

\usepackage{tcolorbox}
\usepackage[frozencache=true,cachedir=.]{minted}
\tcbuselibrary{minted, breakable,xparse,skins}

\definecolor{bg}{gray}{0.95}
\DeclareTCBListing{mintedbox}{O{}m!O{}}{%
  breakable=true,
  listing engine=minted,
  listing only,
  minted language=#2,
  minted style=default,
  minted options={%
    linenos,
    gobble=0,
    breaklines=true,
    breakafter=,,
    fontsize=\small,
    numbersep=8pt,
    #1},
  boxsep=0pt,
  left skip=0pt,
  right skip=0pt,
  left=25pt,
  right=0pt,
  top=3pt,
  bottom=3pt,
  arc=5pt,
  leftrule=0pt,
  rightrule=0pt,
  bottomrule=2pt,
  toprule=2pt,
  colback=bg,
  colframe=orange!70,
  enhanced,
  overlay={%
    \begin{tcbclipinterior}
    \fill[orange!20!white] (frame.south west) rectangle ([xshift=20pt]frame.north west);
    \end{tcbclipinterior}},
  #3}

\ifcolmfinal
\linenumbers
\fi

\title{Self-Generated In-Context Examples Improve LLM Agents \\ for Sequential Decision-Making Tasks}

\author{Vishnu Sarukkai,
Zhiqiang Xie,
Kayvon Fatahalian \\
Stanford University
}

\newcommand{\code}[2][]{\mathsf{\mathtt{#2}_{\text{#1}}}}

\DeclareRobustCommand{\trajbootstrap}{Traj-Bootstrap}
\DeclareRobustCommand{\dbcuration}{+DB-Curation}
\DeclareRobustCommand{\exemplarcuration}{+Exemplar-Curation}
\DeclareRobustCommand{\dbexemplarcuration}{+DB+Exemplar-Curation}

\begin{document}

\maketitle

\begin{abstract}
Improving Large Language Model (LLM) agents for sequential decision-making tasks typically requires extensive task-specific knowledge engineering—custom prompts, curated examples, and specialized observation/action spaces. We investigate a different approach where agents automatically improve by learning from their own successful experiences without human intervention. Our method constructs and refines a database of self-generated trajectories that serve as in-context examples for future tasks. Even naive accumulation of successful trajectories yields substantial performance gains across three diverse benchmarks: ALFWorld (73\% to 89\%), Wordcraft (55\% to 64\%), and InterCode-SQL (75\% to 79\%). These improvements exceed those achieved by upgrading from gpt-4o-mini to gpt-4o and match the performance of allowing multiple attempts per task. We further enhance this approach with two innovations: database-level curation using population-based training to propagate high-performing example collections, and exemplar-level curation that selectively retains trajectories based on their empirical utility as in-context examples. With these enhancements, our method achieves 93\% success on ALFWorld—surpassing approaches that use more powerful LLMs and hand-crafted components. Our trajectory bootstrapping technique demonstrates that agents can autonomously improve through experience, offering a scalable alternative to labor-intensive knowledge engineering.
\end{abstract}

\section{Introduction}
\label{sec:intro}

When creating LLM agents for sequential decision-making tasks, practitioners often improve agent performance by investing in task-specific knowledge engineering (through tedious prompt tuning~\citep{wei2022chain}, human-crafted in-context examples~\citep{brown2020language,wei2023larger} or custom observation and action spaces~\citep{chen2024automanual,yang2024agentoccam}). Using these techniques, scaling agent performance comes from scaling human effort.  

In this paper, we investigate an alternative path: enabling LLM agents to autonomously bootstrap their own performance by leveraging their own successful experiences via in-context learning. The efficacy of in-context learning depends critically on both the quality of the examples~\citep{brown2020language,wei2023larger} and their relevance to the current decision point~\citep{akyurek2022learning,von2023transformers,agarwal2024many}. This insight provides a natural direction for automated agent self-improvement: accumulating successful self-generated trajectories and estimating the most relevant and effective prior experiences to use as in-context examples for each action.

Our work assumes a ReAct-style agent~\citep{yao2023react} that retrieves different examples for each decision point based on their relevance to the current situation \citep{kagaya2024rap, zhou2024trad}. 
We build on this foundation by focusing specifically on how to construct and refine the underlying database of self-generated examples. How can we identify which trajectories enhance performance on new tasks versus those that hinder performance? This database construction problem requires addressing both the collection of high-quality trajectories and the strategic curation of the most valuable ones for future retrieval at each decision point in the agent's reasoning and acting loop. 

We demonstrate that even naive database accumulation improves test-set performance from 73\% to 89\% on ALFWorld, 55\% to 64\% on Wordcraft, and 75\% to 79\% on InterCode-SQL. (Equivalent to what a baseline agent would achieve if it were allowed two to three attempts per task.) We further propose two database construction enhancements: (1) database-level curation that identifies and propagates high-performing example databases, and (2) exemplar-level curation that identifies helpful trajectories based on their empirical utility as in-context examples. These approaches do not require task-specific prompt engineering \citep{fu2024autoguide,chen2024automanual} or custom observation/action space design \citep{zhou2024trad,chen2024automanual}, but improve success rates on ALFWorld to 93\%—surpassing approaches like AutoManual \citep{chen2024automanual} that use more powerful LLMs and hand-crafted observation and action spaces, as well as hierarchical approaches like Autoguide \citep{fu2024autoguide}. The success rate improvement on ALFWorld exceeds the boost obtained from upgrading the agent's underlying LLM from gpt-4o-mini to gpt-4o. Our results highlight the practical value of trajectory bootstrapping as a dimension for scaling test-time compute.

\section{Preliminaries}
\label{sec:preliminaries}

\subsection{Sequential Decision-Making Tasks}

We focus on multi-step sequential decision-making tasks where agents must produce a series of actions over time based on observations of the environment. 
The sequential nature of these tasks introduces unique challenges for LLM agents, as they must interpret intermediate environmental feedback, maintain coherent reasoning across multiple steps, and adapt their strategy based on the evolving task state. This contrasts with one-shot generation tasks (e.g., solving math problems~\citep{hendrycks2021measuring}, one-shot code generation~\citep{jimenez2023swe}) where feedback is only available after the complete solution is provided. 
Our example-driven learning strategy is potentially also suitable to single-step decision-making tasks, but we focus on the multi-step setting due to its applicability to a number of agentic tasks in real-world settings (embodied agents~\citep{song2023llm}, browser-based tasks~\citep{he2024webvoyager}, etc.).

Formally, these tasks can be modeled as Partially Observable Markov Decision Processes (POMDPs), represented by the tuple $(\mathcal{S}, \mathcal{O}, \mathcal{A}, \mathcal{T}, \mathcal{R}, \gamma)$, where $\mathcal{S}$ denotes the underlying state space, $\mathcal{O}$ the observation space, $\mathcal{A}$ the action space, $\mathcal{T}: \mathcal{S} \times \mathcal{A} \rightarrow \mathcal{S}$ defines the deterministic transition function, $\mathcal{R}: \mathcal{S} \times \mathcal{A} \rightarrow \mathbb{R}$ is the reward function, and $\gamma \in [0, 1]$ is the discount factor. The partial observability reflects that agents don't have direct access to the full environment state but rather receive observations that provide limited information.

Given a task goal $g$, an episode consists of the agent interacting with the environment for a maximum of $T$ timesteps. At each timestep $t$, the agent receives an observation $o_t \in \mathcal{O}$ of the current state, takes an action $a_t \in \mathcal{A}$, and the environment transitions to the next state according to the transition function $\mathcal{T}$. In our setting, we specifically consider sparse-reward environments where success is only determined at the end of an episode—the agent receives $\mathcal{R} = 1$ for successful task completion and $\mathcal{R} = 0$ otherwise. This is a standard setting in prior agentic work~\citep{yao2023react,zhao2024expel,fu2024autoguide,chen2024automanual}. 

\subsection{ReAct-style Agent Loop}

Our work assumes a ReAct-style~\citep{yao2023react} agent architecture that employs recent best practices for in-context retrieval~\citep{kagaya2024rap,zhou2024trad}. The agent operates through a three-phase approach (planning, reasoning, and acting) as formalized in Algorithm~\ref{alg:react_stepwise}. Two key components differentiate our implementation from basic ReAct:

\begin{itemize}
    \item To support complex, long-horizon tasks, we incorporate an initial planning step where the agent generates a high-level plan for the entire task before execution begins (~\ref{alg:react_stepwise}, line 3). 
    Full-task planning has been shown to boost agent performance in prior work~\citep{kagaya2024rap,song2023llm}. This modification is fairly standard--either built directly into the algorithm~\citep{kagaya2024rap}, or included in the first reasoning step of human-crafted in-context examples~\citep{zhao2024expel,chen2024automanual}.
    \item Rather than using the same per-task examples throughout an episode~\citep{yao2023react,zhao2024expel,fu2024autoguide}, we follow retrieve different trajectory segments for each decision point, ensuring the agent has access to the most relevant information at each step~\citep{kagaya2024rap,zhou2024trad}. See Appendix~\ref{app:agent_details} for details.
\end{itemize}

\begin{algorithm}[t]
\caption{ReAct-style Agent Loop}
\label{alg:react_stepwise}
\begin{algorithmic}[1]
\Function{Agent}{$g, \mathcal{D}, \mathcal{E}, T$}
    \State \colorbox{gray!20}{$C_p \leftarrow \code{Retrieve}( \mathcal{D},keys=[g])$} \Comment{Retrieve for plan}
    \State $p \leftarrow \code[plan]{LLM}(g, C_p)$ \Comment{Generate initial plan}
    \State Initialize $\tau \leftarrow (g, p, \{\}, -)$ 
    \State $o_1 \leftarrow \mathcal{E}.\code{obs}()$
    \State \colorbox{gray!20}{$C_1 \leftarrow \code{Retrieve}(\mathcal{D},keys=[g,p,o_1])$} \Comment{Retrieve for current observation}
    \For{$t = 1$ to $T$}
        \State $r_t \leftarrow \code[reason]{LLM}(\tau, o_t, C_t)$ \Comment{Generate reasoning}
        \State \colorbox{gray!20}{$C_{t+1} \leftarrow \code{Retrieve}(\mathcal{D},keys=[g,p,r_t])$} \Comment{Retrieve for current reasoning}
        \State $a_t \leftarrow \code[act]{LLM}(\tau, o_t, r_t, C_{t+1})$ \Comment{Decide action}
        \State $o_{t+1}, \text{done}, s \leftarrow \mathcal{E}.\code{step}(a_t)$ \Comment{Execute action in environment}
        \State $\tau \leftarrow \tau \cup (o_t, r_t, a_t)$
        \If{done}
            \State \Return $(g, p, \{(o_i, r_i, a_i)\}_{i=1}^{t}, s)$
        \EndIf
    \EndFor
    \State \Return $(g, p, \{(o_i, r_i, a_i)\}_{i=1}^{T}, 0)$ \Comment{Failed due to timeout}
\EndFunction
\end{algorithmic}
\end{algorithm}

The agent operates through three key LLM-based functions:
\begin{enumerate}
    \item $\code[plan]{LLM}$ generates a high-level plan $p$ for achieving the goal
    \item $\code[reason]{LLM}$ processes the current observation $o_t$ to produce reasoning $r_t$
    \item $\code[act]{LLM}$ determines the appropriate action $a_t$ based on the reasoning
\end{enumerate}

The $\code{Retrieve(·)}$ function selects the $k$ most relevant examples from database $\mathcal{D}$ based on the average cosine distance from the provided lookup keys to the corresponding examples in $\mathcal{D}$--see Appendix~\ref{app:agent_details} and Algorithm~\ref{alg:retrieval_pseudocode} for details. The environment $\mathcal{E}$ provides observations and processes actions, returning the next observation, a termination signal, and the success indicator when the episode ends.

This dynamic retrieval approach is critical for sequential decision-making tasks, as it allows the agent to access specialized knowledge relevant to each unique situation encountered during task execution. Our contribution builds upon this architecture by focusing specifically on how to construct and refine the underlying trajectory database that powers this retrieval mechanism.

Note that Algorithm~\ref{alg:react_stepwise} avoids strategies that employ task-specific prompting, observation spaces~\citep{zhou2024trad} or action spaces~\citep{yang2024agentoccam,chen2024automanual}. The only task-specific knowledge is encapsulated in the content of the trajectory database $\mathcal{D}$.
For simplicity, we eschew other techniques, like hierarchical learning~\citep{zhao2024expel,fu2024autoguide,chen2024automanual}, that are also task-agnostic, but add additional agent complexity. We view the benefits of hierarchical learning as orthogonal and complimentary to our database construction focus.

\section{Problem Statement}
\label{sec:problem_statement}

%\kf{rephrase for a better connection... Given the ReAct-style agent described in Section~\ref{sec:preliminaries}, our goal is to..}
Given the ReAct-style agent described in Section~\ref{sec:preliminaries}, our goal is to construct a trajectory database that maximizes LLM agent performance across sequential decision-making tasks. We focus specifically on how to build and refine the database of examples that the agent retrieves from at each decision point.

Formally, we aim to construct a database $\mathcal{D}$ of trajectories, where each trajectory $\tau \in \mathcal{D}$ captures a complete task attempt:

$$\tau = (g, p, \{(o_t, r_t, a_t)\}_{t=1}^{T}, s)$$

We aim to maximize the agent's expected performance across a distribution of tasks $\mathcal{T}$:

$$\mathcal{D}^* = \arg\max_{\mathcal{D}} \mathbb{E}_{g \sim \mathcal{T}}[\code{Success}(\code{Agent}(g, \mathcal{D}, \mathcal{E}, T))]$$

where $\code{Success(·)}$ returns the binary outcome $s$ of the agent's execution.

We assume that we are given: (1) $\mathcal{D}$ initialized with a small number of human-generated trajectories, (2) a descriptor of the action space, and (3) access to a set of training tasks drawn from $\mathcal{T}$ that the agent can attempt.
All three assumptions are typical in ReAct-based agentic setups~\citep{yao2023react,kagaya2024rap,zhao2024expel,fu2024autoguide,chen2024automanual}. 
Given this setup, our focus is on the agent self-generating trajectories, then choosing the trajectories that should be  added to $\mathcal{D}$ to maximize the agent's expected performance on novel tasks. 

This problem presents two key questions:

\begin{itemize}
    \item \textbf{Scaling Trajectory Collection}: As we gather more successful trajectories, does a larger database provide better guidance for future tasks?
    \item \textbf{Intelligent Trajectory Curation}: How can we determine which subset of collected trajectories will most effectively support the agent in solving new tasks? 
\end{itemize}

\section{Related Work}
\label{sec:related_work}

\paragraph{In-context learning for agent improvement}

Despite the current popularity of reinforcement learning-based approaches for improving agent capabilities~\citep{bai2022training,rafailov2023direct,jaech2024openai,guo2025deepseek}, in-context learning offers distinct scientific and practical advantages. 
These benefits include model-agnostic portability across different LLMs, efficiency in low-sample regimes~\citep{wei2023larger,bertsch2024context}, and accessibility when implementation barriers exist for weight modification methods. Both empirical and theoretical work has established that in-context performance can scale effectively with additional examples~\citep{akyurek2022learning,von2023transformers,bertsch2024context,agarwal2024many}, suggesting that strategic example accumulation should lead to significant performance improvements. We focus on maximizing the value of limited examples through in-context methods, while hypothesizing that database quality, not just quantity, critically influences performance scaling. For completeness, we offer a preliminary investigation in App.~\ref{app:finetune} of how our collected trajectories could potentially serve as training data for fine-tuning approaches.

\paragraph{Automatic in-context examples}

Recent work has demonstrated the effectiveness of optimizing both instructional content and example curation in prompts. DSPy \citep{khattab2023dspy} introduced a framework for optimizing multi-step pipelines through instruction tuning and strategic example curation. Self-generated examples containing reasoning traces can eliminate the need for human-written examples, and these self-generated examples often contribute more to performance than optimized instructions alone~\citep{opsahl2024optimizing}. These approaches typically select fixed exemplars for all task instances, whereas our method enables the dynamic selection of different in-context examples for each decision. 

\paragraph{In-context self-improvement of LLM Agents}

Self-improvement methods for LLM agents either aim to solve one task (performing search/optimization) or transfer knowledge from prior tasks to novel ones (generalization) (see App. ~\ref{app:train_test} for further discussion). Approaches to solve a single task scale the number of sampled solutions at inference time~\citep{brown2024large,wang2024planning,wang2024scaling} or incorporate feedback from failed attempts~\citep{shinn2023reflexion}. Knowledge transfer approaches include abstraction-based methods like ExpeL~\citep{zhao2024expel} and AutoGuide~\citep{fu2024autoguide}, while others employ task-specific information in their design—RAP~\citep{kagaya2024rap} uses task-specific prompts and AutoManual \citep{chen2024automanual} constructs task-specific state and action spaces (see App.~\ref{app:automanual}). Other dimensions of self-improvement include hierarchical execution \citep{wang2023voyager} and optimization techniques for multi-stage systems\citep{saad2024archon,hu2024automated,zhang2024aflow}—techniques complementary to our approach. Rather than developing complex architectures or leveraging task-specific information, we focus on identifying which trajectories most contribute to successful outcomes as in-context examples.

\section{Methods}
\label{sec:methods}

In this section, we discuss three algorithms for 
constructing database $\mathcal{D}$ using a continual collection approach.

\subsection{\emph{\trajbootstrap{}}: Constructing a Database of Previously-Solved Tasks}
\label{sec:basicalgorithm}

Our trajectory-bootstrapping algorithm \trajbootstrap{} constructs a trajectory database $\mathcal{D}$ by collecting successful agent experiences. 
As outlined in Section~\ref{sec:problem_statement}, we start with a minimal set of human-provided exemplars (which could be empty),
then grow the database as the agent successfully completes training tasks. This process creates a positive feedback loop where successful examples help the agent solve new tasks, generating more successful examples.

\trajbootstrap{} operates on principles similar to reward-weighted regression in reinforcement learning~\citep{peters2007reinforcement}, where only successful trajectories ($s = 1$) are stored in the database. 
This filtering mechanism ensures the agent learns from positive examples while avoiding potentially misleading failed attempts. Successful trajectories can be leveraged by asking the agent to imitate the successful patterns in these trajectories. However, failed trajectories are more challenging to operationalize due to the credit attribution problem: it is necessary to identify the `good' vs `bad' parts of the trajectory, before we can even guide the agent to imitate the good parts and avoid the mistakes made in the bad parts. Failed trajectories do offer the opportunity to guide exploration ~\citep{shinn2023reflexion}; we leave this direction to future work.

\subsection{\emph{\dbcuration{}}: Database-Level Data Curation}
\label{sec:db_curation}

\begin{wrapfigure}{r}{0.36\textwidth}  % 'r' for right wrap, adjust width as needed
    \centering
    \vspace{-10pt}
    \includegraphics[width=\linewidth]{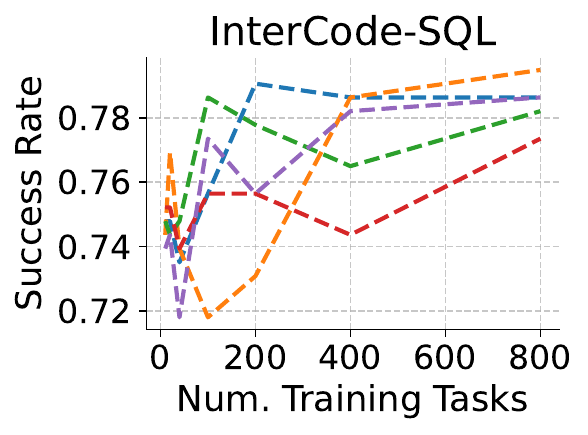}
    \caption{\textbf{\trajbootstrap{} leads to variance in test-time success rate}. Individual trials (5) shown as dashed lines, results on Intercode-SQL benchmark. There is noticeable variability in performance across trials.}
    %\vspace{-10pt}
    \label{fig:db_variance}
\end{wrapfigure}

The \trajbootstrap{} algorithm displays unpredictable performance variation across training trials, even when following identical collection procedures. Figure~\ref{fig:db_variance} illustrates this variation across five trials on the InterCode-SQL benchmark (a benchmark we use for evaluation in Section~\ref{sec:experiments}). 
The variance arises from two factors:
(1) the stochasticity of LLM outputs creating different initial trajectories, and 
(2) an amplification effect where early differences in collected examples lead to wide performance variation.  

This observation motivates a data curation strategy inspired by population-based training in reinforcement learning~\citep{jaderberg2017population}. Figure~\ref{fig:db_variance} shows that some databases lead to better task performance than others—so we identify the underperforming databases periodically during training and remove them, continuing growth from top-performing ones. We introduce \emph{\dbcuration{}}, a population-based training algorithm (Algorithm~\ref{alg:pbt}) to identify and propagate the most effective databases during the bootstrapping process.

\begin{algorithm}[t]
\caption{Database Curation Logic for \dbcuration{}}
\label{alg:pbt}
\begin{algorithmic}[1]
\Procedure{OptimizeDatabases}{$\{\mathcal{D}_1, \mathcal{D}_2, ..., \mathcal{D}_N\}$, $interval$}
    \State Initialize performance metrics $\{m_1, m_2, ..., m_N\}$ for each database
    \For{$t = 1, 2, ..., T_{train}$}
        \For{$i = 1, 2, ..., N$ \textbf{in parallel}}
            \State Execute task $t$ using database $\mathcal{D}_i$
            \State If successful, add trajectory to $\mathcal{D}_i$
            \State Update rolling performance metric $m_i$ on recent tasks
        \EndFor
        \If{$t = 10 \times 2^j$ for any $j \in \mathbb{N}$}
            \State Sort databases by rolling performance on recent tasks
            \State Replace worst database with copies of best
        \EndIf
    \EndFor
\EndProcedure
\end{algorithmic}
\end{algorithm}

We maintain $N$ database instances initialized with identical human-provided exemplars. Each instance is used by a separate agent, accumulating successful trajectories independently. We trigger curation events when the number of tasks attempted reaches size thresholds (starting at size 10 and doubling thereafter: 10, 20, 40, 80, etc.). At each threshold, we evaluate database performance based on the agent's success rate on all training tasks since the last threshold, and we replace the worst-performing database with a copy of the top-performing database.

The key insight of this approach is that database quality emerges from collective properties—like coverage, diversity, and complementarity across examples—not just individual trajectory quality. Moreover, a single trajectory collected early in training can influence many future trajectories by guiding the agent toward particular solution strategies, creating cascading database-level effects. By selecting and propagating entire databases, we preserve these beneficial emergent properties while using a simple, computationally efficient evaluation metric based on recent performance.

\subsection{\emph{\exemplarcuration{}}: Exemplar-Level Data Curation}
\label{sec:exemplar_curation}

While the database-level curation performed by \dbcuration{} identifies entire sets of complementary trajectories, discarding whole databases can eliminate valuable trajectories. We find that even poor-performing databases contain individual high-quality trajectories that yield better outcomes when used as examples. Conversely, some trajectories marked as successful lead to failures when retrieved--because some trajectories lead to success in spite of some bad decisions that would be bad to repeat. 
Explicit identification of high-performing trajectories—those that exemplify generalizable reasoning patterns, as opposed to trajectories containing incorrect reasoning or actions (see Appendix for further analysis)—can improve efficiency compared to wholesale database removal.

This observation motivates \emph{Exemplar-Curation}: identifying and selecting individual high-quality exemplars across multiple database instances based on their empirical utility as in-context examples. This approach parallels value-function learning in reinforcement learning~\citep{barto2021reinforcement}, where we estimate the `value' of each trajectory based on its contribution to successful outcomes.

We introduce a retrieval-weighted quality metric analogous to a value function to quantify each trajectory's contribution to successful outcomes:

\begin{equation}
Q(\tau) = \frac{\sum_{i \in \mathcal{R}(\tau)} o_i \cdot f_i(\tau)}{\sum_{i \in \mathcal{R}(\tau)} f_i(\tau)}
\label{eq:exemplar_metric}
\end{equation}

where $\mathcal{R}(\tau)$ is the set of tasks for which trajectory $\tau$ was retrieved, $o_i$ is the binary outcome of task $i$, and $f_i(\tau)$ is the retrieval frequency during task $i$.

This value metric measures how often a trajectory is associated with successful outcomes when retrieved as an in-context example. It weights outcomes by retrieval frequency, prioritizing trajectories frequently retrieved during successful completions while penalizing those associated with failures. 

\begin{algorithm}[t]
\caption{Database Construction from Top Exemplars for \exemplarcuration{}}
\label{alg:exemplar_curation}
\begin{algorithmic}[1]
\Procedure{SelectExemplars}{$\{\mathcal{D}_1, \mathcal{D}_2, ..., \mathcal{D}_N\}$, $T_{train}$}
    \State $\mathcal{D}_{composite} \gets \emptyset$
    \State Compute quality metric $Q(\tau)$ for each trajectory $\tau \in \bigcup_{i=1}^{N} \mathcal{D}_i$
    \For{each task $t \in T_{train}$}
        \State $T_t \gets$ \{successful trajectories for task $t$ across all databases\}
        \If{$T_t$ is not empty}
            \State Select top-1 trajectory from $T_t$ by quality metric $Q$
            \State Add selected trajectory to $\mathcal{D}_{composite}$
        \EndIf
    \EndFor
    \State \textbf{return} $\mathcal{D}_{composite}$
\EndProcedure
\end{algorithmic}
\end{algorithm}

Algorithm~\ref{alg:exemplar_curation} outlines our exemplar-level curation approach. For each task in the training set, we identify all successful trajectories across database instances and select only the exemplar with the highest value according to the metric. This approach constructs a composite database containing only the most effective exemplars as measured by their empirical contribution to successful outcomes on subsequent tasks.

\subsection{Train-Time vs Test-Time LLM Costs}
\label{sec:train_test_costs}

The curation methods (\dbcuration{} and \exemplarcuration{}) require $N$ times more LLM inference calls during training compared to \trajbootstrap{}, as they maintain $N$ parallel database instances. However, they do not require additional training tasks - all methods use the same task distribution and quantity. At test time, all three methods have identical computational costs--Algorithm~\ref{alg:react_stepwise} is simply provided with a different database $\mathcal{D}$ for each method. This contrasts with approaches that scale the number of LLM calls per test-time task in order to improve performance~\citep{brown2024large,wang2024planning,wang2024scaling,shinn2023reflexion}. Our methods shift computational burden to training while maintaining efficient inference, a property our in-context methods share with fine-tuning methods.

\section{Experiments}
\label{sec:experiments}

We evaluate our database construction methods through experiments addressing three key questions:

\begin{itemize}
    \item Database scaling: How does task success rate scale with increasing database size?
    \item Improving database construction: How much do population-based training and exemplar-level curation improve task success rate? 
    \item Overall effectiveness: How do our approaches compare to alternative approaches leveraging task-specific domain knowledge or hierarchical algorithms? 
\end{itemize}

\subsection{Experimental Setup}

\subsubsection{Benchmark Tasks}

We evaluate our methods on three benchmarks:

\begin{itemize}
    \item \textbf{ALFWorld} \citep{shridhar2020alfworld}: A text-based environment for navigation and object manipulation through textual commands, requiring exploration and sequential reasoning.
    
    \item \textbf{InterCode-SQL} \citep{yang2023intercode}: An interactive coding environment where agents generate SQL queries to answer a user question, requiring understanding of database structures and query semantics.
    
    \item \textbf{Wordcraft} \citep{jiang2020wordcraft}: A simplified adaptation of Little Alchemy, where agents combine elements to create new ones through multi-step processes, requiring compositional reasoning.
\end{itemize}

These benchmarks were selected because they: (1) provide large enough task pools to support meaningful train/test splits, (2) represent diverse reasoning challenges relevant to sequential decision-making, and (3) have been used in prior work, enabling direct comparisons with existing methods. 

\subsubsection{Methods Compared}

Our methods include:
\begin{itemize}
    \item \textbf{Fixed-DB}: The baseline agent as described in Sec.~\ref{sec:preliminaries}, with a fixed database of human-provided initial examples and no database growth.

    \item \textbf{\trajbootstrap{}}: The simple progressive accumulation approach from Sec.~\ref{sec:basicalgorithm}. %A simple database construction approach that progressively accumulates successful trajectories into a growing database. Each task in the training set of a given benchmark is attempted once, generating a single trajectory that is added to the database if successful. \kf{cut this down. All you need to say is the simple progressive accumulation approach from 5.1. Can be a short sentence.}
    
    \item \textbf{\trajbootstrap{}\dbcuration{}}: Our database-level trajectory curation from Alg.~\ref{alg:pbt}. %to maintain multiple trajectory databases and selectively propagate the ones with the highest success rates.

    \item \textbf{\trajbootstrap{}\exemplarcuration{}}: Our exemplar-level trajectory curation from Alg.~\ref{alg:exemplar_curation}.% to identify and preserve individual high-quality trajectories based on their empirical contribution to successful outcomes.

    \item \textbf{\trajbootstrap{}\dbexemplarcuration{}}: Applying both our database-level trajectory curation and propagation and our exemplar-level trajectory curation. 
\end{itemize}

We compare these methods to two more advanced hierarchical designs. \textbf{Autoguide} \citep{fu2024autoguide} converts successful trajectories into explicit rules and retrieves the most contextually relevant rules, alongside low-level trajectories, at inference time. \textbf{AutoManual} \citep{chen2024automanual} leverages hand-crafted task-specific observation and action spaces--see Appendix~\ref{app:automanual} for details.

\subsubsection{Implementation Details}

Unless otherwise specified, we use GPT-4o-mini as our base LLM (temperature 0.1).
For Fixed-DB and all \trajbootstrap{} agents, we retrieve the top-$k$ most similar trajectories at each decision step ($k=6$ for ALFWorld and InterCode-SQL, $10$ for Wordcraft).
We initialize each database with a small human-provided example set ($18$ for ALFWorld, $10$ for InterCode-SQL, $4$ for Wordcraft). 
With \dbcuration{}, we maintain five database instances with curation every time the database size is doubled, starting with a minimum size of ten trajectories. 
We report success rates averaged over five random seeds. 
By default, we report success rate given the database at the end of the training process.  

\subsection{\trajbootstrap{} Results}
\label{sec:results}

\paragraph{\trajbootstrap{} performance improves with more training tasks}
\label{sec:results_db_scaling}

\begin{table}[t]
    \centering
    \begin{tabularx}{0.85\textwidth}{l r r r}
    \toprule
    \textbf{Method} & \textbf{ALFWorld} & \textbf{InterCode-SQL} & \textbf{Wordcraft} \\
    \midrule
    Fixed-DB & 0.73$\pm$0.02 & 0.75$\pm$0.01 & 0.55$\pm$0.03 \\
    \trajbootstrap{} & 0.89$\pm$0.01 & 0.79$\pm$0.01 & 0.64$\pm$0.03 \\
    \quad\dbcuration{} & 0.91$\pm$0.01 & 0.78$\pm$0.01 & 0.64$\pm$0.01 \\
    \quad\exemplarcuration{} & 0.90$\pm$0.02 & 0.81$\pm$0.01 & \textbf{0.72}$\pm$0.02 \\
    \quad\dbexemplarcuration{} & \textbf{0.93}$\pm$0.03 & \textbf{0.82}$\pm$0.01 & 0.69$\pm$0.01 \\
    \bottomrule
    \end{tabularx}
    \caption{\textbf{Average success rate of our methods: self-collected trajectories provide the largest boosts in task success rate}. \trajbootstrap{} outperforms Fixed-DB across all three benchmarks. The combination of \dbcuration{} and \exemplarcuration{} provides the best performance on both ALFWorld and InterCode-SQL. \exemplarcuration{} provides the best performance on Wordcraft.}
    \vspace{-1.0em}
    \label{tab:ours_all_benchmarks}
\end{table}

\begin{table}[t]
    \centering
    \begin{tabularx}{0.86\textwidth}{l l r r}
    \toprule
    \textbf{Method} & \textbf{LLM(s)} & \textbf{Num Training Tasks} & \textbf{ALFWorld} \\
    \midrule
    Autoguide~\citep{fu2024autoguide} & \makecell[l]{gpt-3.5-turbo\\+ gpt-4-turbo} & 100 & 0.79* \\
    \midrule
    \multirow{2}{*}{Automanual~\citep{chen2024automanual}} & gpt-4o-mini & 36 & 0.72$\pm$0.01 \\
    & \makecell[l]{gpt-4-turbo\\+ gpt-4o-mini} & 36 & 0.91$\pm$0.01 \\
    \midrule
    \multirow{2}{*}{Fixed-DB} & gpt-4o-mini & 0 & 0.73$\pm$0.05 \\
     & gpt-4o & 0 & 0.88$\pm$0.02 \\
    \midrule
    \multirow{2}{*}{\trajbootstrap{}} & gpt-4o-mini & 100 & 0.84$\pm$0.04 \\
                                   & gpt-4o-mini & 3500 & 0.89$\pm$0.01 \\
    \midrule
    \multirow{2}{*}{\dbcuration{}} & gpt-4o-mini & 100 & 0.86$\pm$0.02 \\
                             & gpt-4o-mini & 3500 & 0.91$\pm$0.01 \\
    \midrule
    \multirow{2}{*}{\exemplarcuration{}} & gpt-4o-mini & 100 & 0.86$\pm$0.03 \\
                             & gpt-4o-mini & 3500 & 0.90$\pm$0.02 \\
    \midrule
    \multirow{2}{*}{\makecell[l]{\dbcuration{} \\ \exemplarcuration{}}} & gpt-4o-mini & 100 & 0.81$\pm$0.02 \\
                             & gpt-4o-mini & 3500 & \textbf{0.93}$\pm$0.03 \\
    \bottomrule
    \end{tabularx}
    \caption{ 
    \textbf{Comparison of agent success rates on ALFWorld: contextualizing the performance of Traj-Boostrap.}
    The 15-point boost in average success rate from database construction via \trajbootstrap{} is similar to that achieved from upgrading Fixed-DB from gpt-4o-mini to gpt-4o. The performance of \trajbootstrap{}\dbexemplarcuration{} exceeds Automanual~\citep{chen2024automanual}, even though Automanual utilizes hand-designed observation and action spaces and a better LLM (gpt-4-turbo+gpt-4o-mini). * indicates results reported from original papers. 
    }
    \vspace{-1.0em}
    \label{tab:alfworld_results}
\end{table}

\begin{figure}[t]
    \centering
    \begin{minipage}{\textwidth}
        \centering
        \begin{subfigure}[t]{0.32\textwidth}
            \centering
            \includegraphics[width=\textwidth]{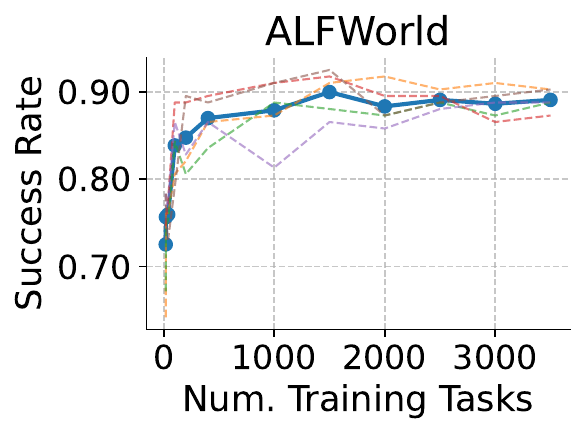}
            \label{fig:alfworld_results}
        \end{subfigure}
        \hfill
        \begin{subfigure}[t]{0.32\textwidth}
            \centering
            \includegraphics[width=\textwidth]{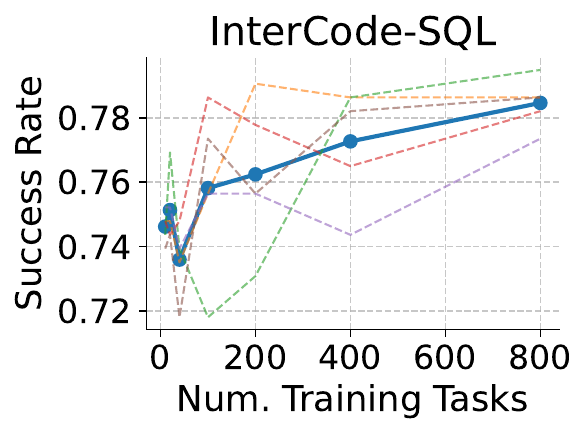}
            \label{fig:intercode_results}
        \end{subfigure}
        \hfill
        \begin{subfigure}[t]{0.32\textwidth}
            \centering
            \includegraphics[width=\textwidth]{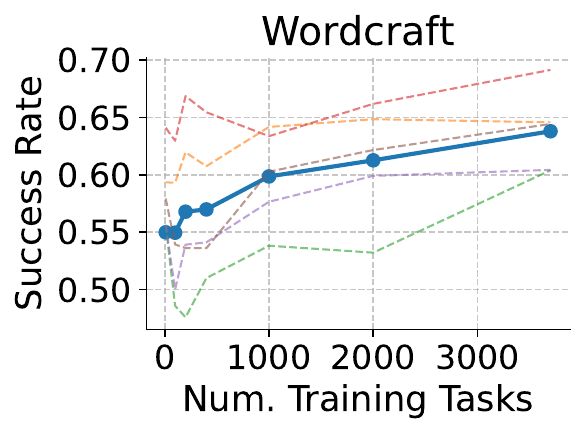}
            \label{fig:wordcraft_results}
        \end{subfigure}
    \end{minipage}

    \vspace{-1.5em} % <-- This now works correctly

    \begin{minipage}{\textwidth}
        \centering
        \includegraphics[width=0.2\textwidth]{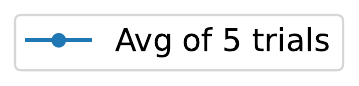}
    \end{minipage}

    \vspace{-1.0em}

    \caption{\textbf{\trajbootstrap{} results: success rate improves with increasing training tasks on all three benchmarks}. Individual trials (5) shown as dashed lines. All benchmarks exhibit diminishing returns as the database size increases. Trials show substantial performance variability, both within individual trials and across different trials.}
    \vspace{-1.0em}
    \label{fig:db_scaling}
\end{figure}

Tab.~\ref{tab:ours_all_benchmarks} presents the final success rate metrics for our database construction methods. The performance of \trajbootstrap{} generally improves with increases in the number of training tasks attempted (Fig.~\ref{fig:db_scaling}) 
%On ALFWorld, success rate increases from 0.73 to 0.89 as the database grows from 18 initial examples to 3500 accumulated trajectories. InterCode-SQL improves from 0.75 to 0.79 and Wordcraft from 0.55 to 0.64. \kf{clear in the table, don't think you need the preceding text with all the numbers.} 
Performance continues to improve with more training tasks across all benchmarks, but exhibits diminishing returns—most gains occur within the first 25\% of added training tasks. This efficiency decline occurs because each new example is retrieved less frequently as the database grows, influencing fewer generations, a pattern consistent with findings from \citet{bertsch2024context} and \citet{agarwal2024many}.
As mentioned in Sec.~\ref{sec:methods}, we observe performance variability across trials and within individual trials. Cross-trial variance indicates that some trials produce higher-performing databases when solving identical tasks. Within-trial fluctuations show that certain added trajectories can degrade performance. %These patterns highlight the potential value of \dbcuration{} and \exemplarcuration{}.

\paragraph{\dbcuration{} boosts performance on ALFWorld}
\label{sec:results_multi_trial}

\begin{figure}[t]
    \centering
    \begin{minipage}{\textwidth}
        \centering
        \begin{subfigure}[b]{0.32\textwidth}
            \centering
            \includegraphics[width=\textwidth]{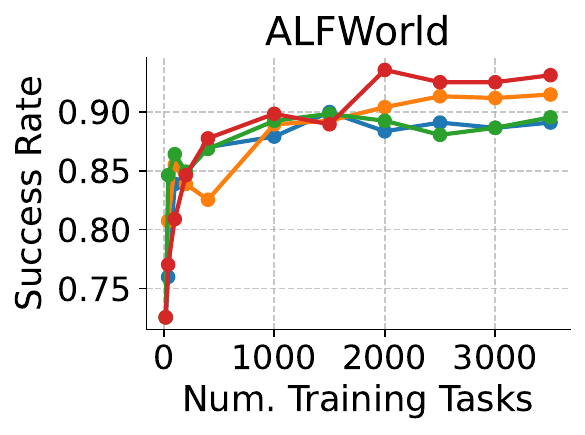}
            %\caption{ALFWorld}
            \label{fig:advanced_alfworld}
        \end{subfigure}
        \hfill
        \begin{subfigure}[b]{0.32\textwidth}
            \centering
            \includegraphics[width=\textwidth]{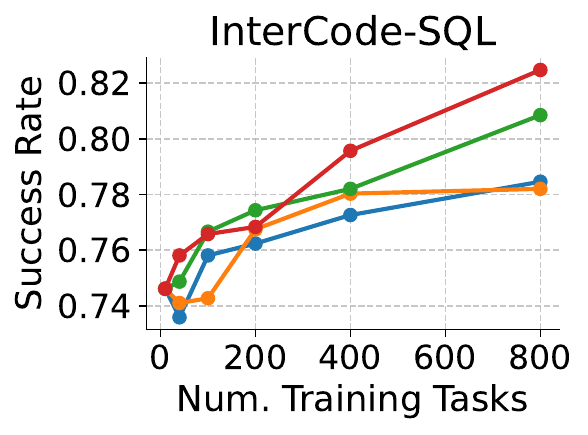}
            %\caption{InterCode-SQL}
            \label{fig:advanced_intercode}
        \end{subfigure}
        \hfill
        \begin{subfigure}[b]{0.32\textwidth}
            \centering
            \includegraphics[width=\textwidth]{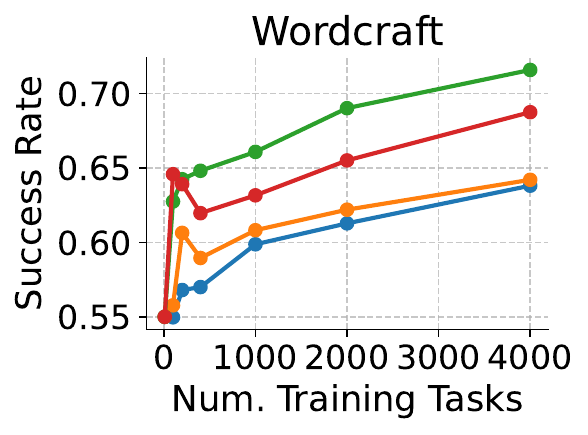}
            %\caption{Wordcraft}
            \label{fig:advanced_wordcraft}
        \end{subfigure}
    \end{minipage}

    \vspace{-1.5em} % adjust this value to control spacing between graphs and legend

    \begin{minipage}{\textwidth}
        \centering
        \includegraphics[width=0.95\textwidth]{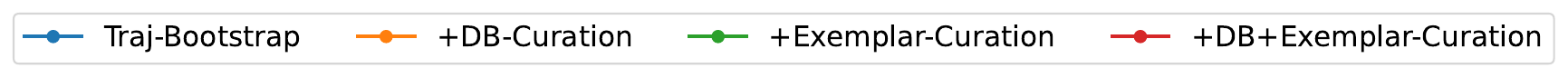}
    \end{minipage}

    \vspace{-1.0em}

    \caption{\textbf{Success rate comparison for \trajbootstrap{} and its variants (\dbcuration{}, \exemplarcuration{}, \dbexemplarcuration{}).} 
    \dbcuration{} enhances final success rate only on ALFWorld, but improves success rate for smaller DB sizes on all benchmarks. \exemplarcuration{} delivers success rate gains on both Intercode-SQL and Wordcraft. The combination of both enhancements delivers the largest gains on both ALFWorld and InterCode-SQL.}
    \label{fig:multitrial_results}
\end{figure}

Fig.~\ref{fig:multitrial_results} illustrates how \dbcuration{} can improve upon \trajbootstrap{}'s performance, despite exhibiting occasional performance dips at smaller database sizes. These dips result from inaccurate estimates (due to low sample count) of database quality early in the process—introducing noise into the curation process. 
%As discussed in Sec.~\ref{sec:train_test_costs}, \dbcuration{} incurs additional train-time LLM costs to construct multiple independent databases--but requires neither extra training tasks nor additional test-time costs. 

\paragraph{\exemplarcuration{} boosts performance on InterCode-SQL and Wordcraft}
\label{sec:results_filtering}

\begin{figure}[t]
    \centering
    \begin{minipage}{\textwidth}
        \centering
        \begin{subfigure}[b]{0.32\textwidth}
            \centering
            \includegraphics[width=\textwidth]{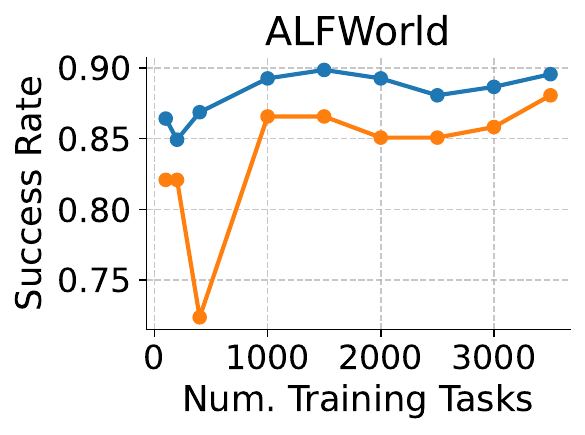}
            %\caption{ALFWorld}
            \label{fig:best_worst_alfworld}
        \end{subfigure}
        \hfill
        \begin{subfigure}[b]{0.32\textwidth}
            \centering
            \includegraphics[width=\textwidth]{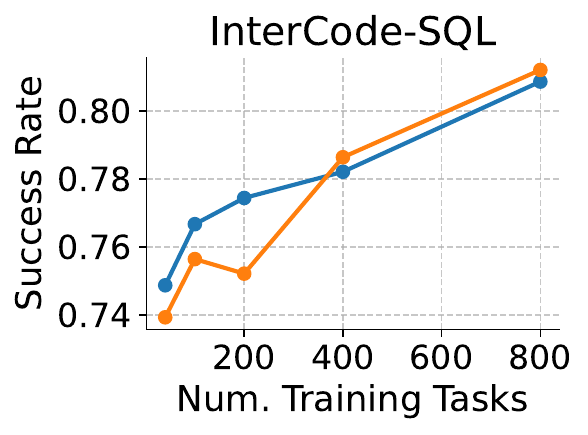}
            %\caption{InterCode-SQL}
            \label{fig:best_worst_intercode}
        \end{subfigure}
        \hfill
        \begin{subfigure}[b]{0.32\textwidth}
            \centering
            \includegraphics[width=\textwidth]{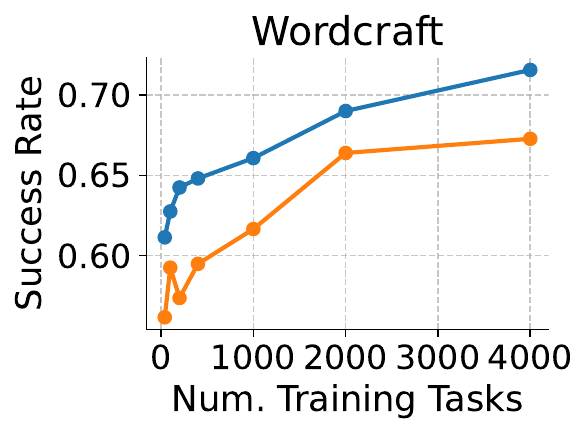}
            %\caption{Wordcraft}
            \label{fig:best_worst_wordcraft}
        \end{subfigure}
    \end{minipage}

    \vspace{-1.5em} % adjust this for spacing between graphs and legend

    \begin{minipage}{\textwidth}
        \centering
        \includegraphics[width=0.25\textwidth]{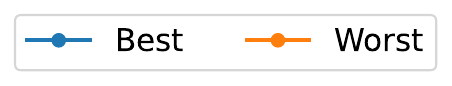}
    \end{minipage}
    \vspace{-1.0em}

    \caption{\textbf{The `best' bootstrapped trajectories compared to the `worst'}. 
    Databases constructed from the highest-quality successful trajectory per task, as measured by Eq.~\ref{eq:exemplar_metric}, outperform databases built from the lowest-quality successful trajectories on both ALFWorld and Wordcraft. The `best' curve is identical to \exemplarcuration{}, while the `worst' curve selects the bottom-1 trajectory instead of top-1 in Alg.~\ref{alg:exemplar_curation}, line 7.}
    \label{fig:best_worst}
    \vspace{-1.0em}
\end{figure}

As seen in Tab.~\ref{tab:ours_all_benchmarks}, \exemplarcuration{} yields improvements in final task success rates on InterCode-SQL and Wordcraft, and also boosts success rate at intermediate database sizes for both InterCode-SQL and Wordcraft (Fig.\ref{fig:multitrial_results}).
To further highlight the impact of our exemplar-level curation metric, 
Fig.~\ref{fig:best_worst} compares databases built from the `best' trajectories that are the most empirically effective in-context examples versus the least effective trajectories, as determined by Equation~\ref{eq:exemplar_metric} in Sec.~\ref{sec:exemplar_curation}. The `best' curve is identical to \exemplarcuration{}, while the `worst' curve selects the bottom-1 trajectory instead of top-1 in Alg.~\ref{alg:exemplar_curation}, line 7. Using the database of high-quality examples yields a higher success rate across all database sizes for ALFWorld and Wordcraft, and for smaller database sizes for InterCode-SQL. 
%\kf{the lack of magnitude of this effect on 2/3 is a little surprising. Anything we should comment on here?  This text makes it read like a big win.}

\paragraph{\dbexemplarcuration{} achieves best performance on ALFWorld and InterCode-SQL}
Fig.~\ref{fig:multitrial_results} and Tab.~\ref{tab:ours_all_benchmarks} highlight that \dbcuration{} and \exemplarcuration{} can be complementary, as the combined \dbexemplarcuration{} achieves the best final task success rates on both ALFWorld and InterCode-SQL (0.93 and 0.82 respectively). Note that on Wordcraft, \dbcuration{} fails to provide boosts whether or not \exemplarcuration{} is used--\trajbootstrap{} and \dbcuration{} perform identically (0.64), and \exemplarcuration{}(0.71) outperforms \dbexemplarcuration{} (0.67). 

\subsection{Contextualizing performance boosts from \trajbootstrap{}}
\vspace{-0.5em}

To contextualize the improvements achieved by \trajbootstrap{}, we compare with several alternative strategies: test-time sampling, using a better LLM, task-specific strategies, and hierarchical strategies. 
\vspace{-0.5em}

\paragraph{Comparison with test-time scaling}
\label{sec:test_time}

\begin{table}[t]
\centering
\begin{tabular}{lrrr}
\toprule
\textbf{Method} & \textbf{ALFWorld} & \textbf{Intercode-SQL} & \textbf{Wordcraft} \\
\trajbootstrap{} & 0.89$\pm$0.01 & 0.79$\pm$0.01 & 0.64$\pm$0.03 \\
\quad\dbexemplarcuration{} & 0.93$\pm$0.03 & 0.82$\pm$0.01 & 0.69$\pm$0.01 \\
\midrule
Fixed-DB@1 & 0.73$\pm$0.03 & 0.75$\pm$0.01 & 0.55$\pm$0.03 \\
Fixed-DB@2 & 0.87$\pm$0.02 & 0.78$\pm$0.03 & 0.62$\pm$0.02 \\
Fixed-DB@3 & 0.92$\pm$0.02 & 0.80$\pm$0.03 & 0.64$\pm$0.02 \\
Fixed-DB@4 & 0.94$\pm$0.02 & 0.81$\pm$0.02 & 0.66$\pm$0.02 \\
Fixed-DB@5 & 0.96$\pm$0.02 & 0.82$\pm$0.03 & 0.72$\pm$0.02 \\
\bottomrule
\end{tabular}
\caption{\textbf{Pass@k of Fixed-DB on all benchmarks.} On all benchmarks, using only a single test-time attempt per task, \trajbootstrap{} achieves success rates between that of Fixed-DB at pass@2 and at pass@3. \dbexemplarcuration{} achieves success rates between pass@3 and pass@5. 
}
\vspace{-1.5em}
\label{tab:pass_at_k}
\end{table}

Our trajectory bootstrapping approaches achieve success rate improvements equivalent to scaling test-time compute by making multiple task attempts—an advantage in scenarios where multiple attempts are impractical or when success verification is unavailable at test time. Furthermore, our approaches provide these benefits without requiring any modifications to the test-time inference process. To demonstrate the magnitude of this benefit, we compare our methods to the alternative strategy of making multiple attempts at each test task with the Fixed-DB baseline and selecting the best outcome~\citep{brown2024large,wang2024planning,wang2024scaling}.
Tab.~\ref{tab:pass_at_k} reports the pass@k metrics for Fixed-DB across all three benchmarks, representing the probability of at least one successful attempt when making $k$ independent attempts at each task. 
Using only a single attempt per task, \trajbootstrap{} approach achieves success rate comparable to Fixed-DB pass@2 or pass@3 on all three benchmarks.
\dbcuration{} and/or \exemplarcuration{} perform nearly on the level of Fixed-DB pass@4 on ALFWorld, pass@5 on InterCode-SQL, and pass@5 for Wordcraft. 

\paragraph{Comparsion with model upgrades}
\label{sec:better_model}
\vspace{-0em}

On ALFWorld, after 3500 training tasks \trajbootstrap{} yields a 20-point success rate boost over Fixed-DB, significantly outperforming the 15-point improvement gained by upgrading Fixed-DB to a more powerful LLM. 
%\kf{could mention better results that get into 90's, saying it significantly outperforms the LLM upgrade. Also leads into next section well.}
%\kf{drop next sentence:} For context, Fixed-DB with GPT-4o-mini only achieves 0.73, highlighting the boost provided by either database construction or model scaling.
\vspace{-0em}

\paragraph{Comparison with task-specific strategies}
\label{sec:results_comparison}
Tab.~\ref{tab:alfworld_results} shows that \trajbootstrap{}\dbexemplarcuration{} using GPT-4o-mini achieves a success rate exceeds (0.93) that exceeds that of Automanual~\citep{chen2024automanual} configured to use a  combination of GPT-4-turbo and GPT-4o-mini (0.91). Thus, our methods outperform an approach that uses a more powerful LLM and customized observation and action spaces.
See App.~\ref{app:intercode_handcrafted} for a comparison to hand-crafted approaches on InterCode-SQL.
\vspace{-0.5em}

\paragraph{Comparison with hierarchical algorithms}
Given $100$ training tasks, Autoguide, a hierarchical rule-learning approach, achieves a 0.79 success rate (using a combination of gpt-3.5-turbo + gpt-4-turbo). Given the same number of training tasks our best approach achieves significantly greater success rate (0.86) with gpt-4o-mini (Tab.~\ref{tab:alfworld_results}). While this comparison employs different algorithms and LLMs, the performance of \trajbootstrap{} suggests that self-constructed databases of low-level trajectories can be competitive with hierarchical approaches. 
\vspace{-0.5em}

\subsection{Extending \trajbootstrap{}}
\vspace{-0.5em}

\paragraph{Can we predict agent success?} Beyond improving agent performance, we can also utilize our self-collected examples to implement useful agent diagnostics, such as predicting an agent's success on novel tasks. On InterCode-SQL and Wordcraft, we train a calibrated Random Forest classifier of agent success based on task goal and initial observation embeddings. Classifier quality (measured via AUROC) improves with database size, reaching 0.77 for InterCode-SQL and 0.71 for Wordcraft with our largest databases. The predicted probabilities also closely match empirical success rates, indicating well-calibrated predictions. See App.~\ref{app:prediction} for details.
\vspace{-0.5em}

\paragraph{Can we use our self-collected databases for fine-tuning?} We fine-tune GPT-4o-mini using trajectories from our best-performing \trajbootstrap{}\dbexemplarcuration{} database for each benchmark. The resulting fine-tuned agents (ReAct-Finetune) outperform our in-context approach on ALFWorld (23-point vs. 20-point boost) and Wordcraft (19-point vs. 14-point), but perform worse on InterCode-SQL (4-point vs. 7-point). 
%\kf{it might be more helpful to express these in the text as boosts in success rate e.g. 4 vs. 2. The result is that a point boost allows me to simulatenously assess magnitude of benefit and compare the two methods.} 
This suggests our self-collected examples are effective not only for in-context learning but also for creating fine-tuned agents. See App.~\ref{app:finetune} for details.
\vspace{-0.5em}

\section{Discussion}
\label{sec:discussion}

The success of our approach reveals performance gains that stem primarily from accumulating successful examples, establishing a foundation for agent self-improvement where the quantity and quality of accessible data rivals the importance of architectural complexity. This parallels trends in traditional deep learning, where data curation often yields substantial improvements. Our findings point to promising research directions that approach LLM agent enhancement from a data-centric perspective—advancing both strategic data collection methods (balancing exploration versus exploitation across diverse tasks) and refined filtering techniques to maximize performance.

\paragraph{Acknowledgments}
Thank you to Brennan Shacklett, Purvi Goel, Zander Majercik, William Wang, Bradley Brown, Jon Saad-Falcon, and William Mark for valuable discussions and feedback. Support for this project was provided by Roblox and Meta, and API credits were provided by OpenAI and together.ai. 

\bibliography{colm2025_conference}

\begin{thebibliography}{42}
\providecommand{\natexlab}[1]{#1}
\providecommand{\url}[1]{\texttt{#1}}
\expandafter\ifx\csname urlstyle\endcsname\relax
  \providecommand{\doi}[1]{doi: #1}\else
  \providecommand{\doi}{doi: \begingroup \urlstyle{rm}\Url}\fi

\bibitem[Wei et~al.(2022)Wei, Wang, Schuurmans, Bosma, Xia, Chi, Le, Zhou, et~al.]{wei2022chain}
Jason Wei, Xuezhi Wang, Dale Schuurmans, Maarten Bosma, Fei Xia, Ed~Chi, Quoc~V Le, Denny Zhou, et~al.
\newblock Chain-of-thought prompting elicits reasoning in large language models.
\newblock \emph{Advances in neural information processing systems}, 35:\penalty0 24824--24837, 2022.

\bibitem[Brown et~al.(2020)Brown, Mann, Ryder, Subbiah, Kaplan, Dhariwal, Neelakantan, Shyam, Sastry, Askell, et~al.]{brown2020language}
Tom Brown, Benjamin Mann, Nick Ryder, Melanie Subbiah, Jared~D Kaplan, Prafulla Dhariwal, Arvind Neelakantan, Pranav Shyam, Girish Sastry, Amanda Askell, et~al.
\newblock Language models are few-shot learners.
\newblock \emph{Advances in neural information processing systems}, 33:\penalty0 1877--1901, 2020.

\bibitem[Wei et~al.(2023)Wei, Wei, Tay, Tran, Webson, Lu, Chen, Liu, Huang, Zhou, et~al.]{wei2023larger}
Jerry Wei, Jason Wei, Yi~Tay, Dustin Tran, Albert Webson, Yifeng Lu, Xinyun Chen, Hanxiao Liu, Da~Huang, Denny Zhou, et~al.
\newblock Larger language models do in-context learning differently.
\newblock \emph{arXiv preprint arXiv:2303.03846}, 2023.

\bibitem[Chen et~al.(2024)Chen, Li, Yang, Yu, Lin, and He]{chen2024automanual}
Minghao Chen, Yihang Li, Yanting Yang, Shiyu Yu, Binbin Lin, and Xiaofei He.
\newblock Automanual: Generating instruction manuals by llm agents via interactive environmental learning.
\newblock \emph{arXiv preprint arXiv:2405.16247}, 2024.

\bibitem[Yang et~al.(2024)Yang, Liu, Chaudhary, Fakoor, Chaudhari, Karypis, and Rangwala]{yang2024agentoccam}
Ke~Yang, Yao Liu, Sapana Chaudhary, Rasool Fakoor, Pratik Chaudhari, George Karypis, and Huzefa Rangwala.
\newblock Agentoccam: A simple yet strong baseline for llm-based web agents.
\newblock \emph{arXiv preprint arXiv:2410.13825}, 2024.

\bibitem[Aky{\"u}rek et~al.(2022)Aky{\"u}rek, Schuurmans, Andreas, Ma, and Zhou]{akyurek2022learning}
Ekin Aky{\"u}rek, Dale Schuurmans, Jacob Andreas, Tengyu Ma, and Denny Zhou.
\newblock What learning algorithm is in-context learning? investigations with linear models.
\newblock \emph{arXiv preprint arXiv:2211.15661}, 2022.

\bibitem[Von~Oswald et~al.(2023)Von~Oswald, Niklasson, Randazzo, Sacramento, Mordvintsev, Zhmoginov, and Vladymyrov]{von2023transformers}
Johannes Von~Oswald, Eyvind Niklasson, Ettore Randazzo, Jo{\~a}o Sacramento, Alexander Mordvintsev, Andrey Zhmoginov, and Max Vladymyrov.
\newblock Transformers learn in-context by gradient descent.
\newblock In \emph{International Conference on Machine Learning}, pages 35151--35174. PMLR, 2023.

\bibitem[Agarwal et~al.(2024)Agarwal, Singh, Zhang, Bohnet, Rosias, Chan, Zhang, Anand, Abbas, Nova, et~al.]{agarwal2024many}
Rishabh Agarwal, Avi Singh, Lei Zhang, Bernd Bohnet, Luis Rosias, Stephanie Chan, Biao Zhang, Ankesh Anand, Zaheer Abbas, Azade Nova, et~al.
\newblock Many-shot in-context learning.
\newblock \emph{Advances in Neural Information Processing Systems}, 37:\penalty0 76930--76966, 2024.

\bibitem[Yao et~al.(2023)Yao, Zhao, Yu, Du, Shafran, Narasimhan, and Cao]{yao2023react}
Shunyu Yao, Jeffrey Zhao, Dian Yu, Nan Du, Izhak Shafran, Karthik Narasimhan, and Yuan Cao.
\newblock React: Synergizing reasoning and acting in language models.
\newblock In \emph{International Conference on Learning Representations (ICLR)}, 2023.

\bibitem[Kagaya et~al.(2024)Kagaya, Yuan, Lou, Karlekar, Pranata, Kinose, Oguri, Wick, and You]{kagaya2024rap}
Tomoyuki Kagaya, Thong~Jing Yuan, Yuxuan Lou, Jayashree Karlekar, Sugiri Pranata, Akira Kinose, Koki Oguri, Felix Wick, and Yang You.
\newblock Rap: Retrieval-augmented planning with contextual memory for multimodal llm agents.
\newblock \emph{arXiv preprint arXiv:2402.03610}, 2024.

\bibitem[Zhou et~al.(2024)Zhou, Yang, Wen, Wen, Wang, Xi, Xu, Yu, and Zhang]{zhou2024trad}
Ruiwen Zhou, Yingxuan Yang, Muning Wen, Ying Wen, Wenhao Wang, Chunling Xi, Guoqiang Xu, Yong Yu, and Weinan Zhang.
\newblock Trad: Enhancing llm agents with step-wise thought retrieval and aligned decision.
\newblock In \emph{Proceedings of the 47th International ACM SIGIR Conference on Research and Development in Information Retrieval}, pages 3--13, 2024.

\bibitem[Fu et~al.(2024)Fu, Kim, Kim, Sohn, Logeswaran, Bae, and Lee]{fu2024autoguide}
Yao Fu, Dong-Ki Kim, Jaekyeom Kim, Sungryull Sohn, Lajanugen Logeswaran, Kyunghoon Bae, and Honglak Lee.
\newblock Autoguide: Automated generation and selection of context-aware guidelines for large language model agents.
\newblock \emph{arXiv preprint arXiv:2403.08978}, 2024.

\bibitem[Hendrycks et~al.(2021)Hendrycks, Burns, Kadavath, Arora, Basart, Tang, Song, and Steinhardt]{hendrycks2021measuring}
Dan Hendrycks, Collin Burns, Saurav Kadavath, Akul Arora, Steven Basart, Eric Tang, Dawn Song, and Jacob Steinhardt.
\newblock Measuring mathematical problem solving with the math dataset.
\newblock \emph{arXiv preprint arXiv:2103.03874}, 2021.

\bibitem[Jimenez et~al.(2023)Jimenez, Yang, Wettig, Yao, Pei, Press, and Narasimhan]{jimenez2023swe}
Carlos~E Jimenez, John Yang, Alexander Wettig, Shunyu Yao, Kexin Pei, Ofir Press, and Karthik Narasimhan.
\newblock Swe-bench: Can language models resolve real-world github issues?
\newblock \emph{arXiv preprint arXiv:2310.06770}, 2023.

\bibitem[Song et~al.(2023)Song, Wu, Washington, Sadler, Chao, and Su]{song2023llm}
Chan~Hee Song, Jiaman Wu, Clayton Washington, Brian~M Sadler, Wei-Lun Chao, and Yu~Su.
\newblock Llm-planner: Few-shot grounded planning for embodied agents with large language models.
\newblock In \emph{Proceedings of the IEEE/CVF international conference on computer vision}, pages 2998--3009, 2023.

\bibitem[He et~al.(2024)He, Yao, Ma, Yu, Dai, Zhang, Lan, and Yu]{he2024webvoyager}
Hongliang He, Wenlin Yao, Kaixin Ma, Wenhao Yu, Yong Dai, Hongming Zhang, Zhenzhong Lan, and Dong Yu.
\newblock Webvoyager: Building an end-to-end web agent with large multimodal models.
\newblock \emph{arXiv preprint arXiv:2401.13919}, 2024.

\bibitem[Zhao et~al.(2024)Zhao, Huang, Xu, Lin, Liu, and Huang]{zhao2024expel}
Andrew Zhao, Daniel Huang, Quentin Xu, Matthieu Lin, Yong-Jin Liu, and Gao Huang.
\newblock Expel: Llm agents are experiential learners.
\newblock In \emph{Proceedings of the AAAI Conference on Artificial Intelligence}, volume~38, pages 19632--19642, 2024.

\bibitem[Bai et~al.(2022)Bai, Jones, Ndousse, Askell, Chen, DasSarma, Drain, Fort, Ganguli, Henighan, et~al.]{bai2022training}
Yuntao Bai, Andy Jones, Kamal Ndousse, Amanda Askell, Anna Chen, Nova DasSarma, Dawn Drain, Stanislav Fort, Deep Ganguli, Tom Henighan, et~al.
\newblock Training a helpful and harmless assistant with reinforcement learning from human feedback.
\newblock \emph{arXiv preprint arXiv:2204.05862}, 2022.

\bibitem[Rafailov et~al.(2023)Rafailov, Sharma, Mitchell, Manning, Ermon, and Finn]{rafailov2023direct}
Rafael Rafailov, Archit Sharma, Eric Mitchell, Christopher~D Manning, Stefano Ermon, and Chelsea Finn.
\newblock Direct preference optimization: Your language model is secretly a reward model.
\newblock \emph{Advances in Neural Information Processing Systems}, 36:\penalty0 53728--53741, 2023.

\bibitem[Jaech et~al.(2024)Jaech, Kalai, Lerer, Richardson, El-Kishky, Low, Helyar, Madry, Beutel, Carney, et~al.]{jaech2024openai}
Aaron Jaech, Adam Kalai, Adam Lerer, Adam Richardson, Ahmed El-Kishky, Aiden Low, Alec Helyar, Aleksander Madry, Alex Beutel, Alex Carney, et~al.
\newblock Openai o1 system card.
\newblock \emph{arXiv preprint arXiv:2412.16720}, 2024.

\bibitem[Guo et~al.(2025)Guo, Yang, Zhang, Song, Zhang, Xu, Zhu, Ma, Wang, Bi, et~al.]{guo2025deepseek}
Daya Guo, Dejian Yang, Haowei Zhang, Junxiao Song, Ruoyu Zhang, Runxin Xu, Qihao Zhu, Shirong Ma, Peiyi Wang, Xiao Bi, et~al.
\newblock Deepseek-r1: Incentivizing reasoning capability in llms via reinforcement learning.
\newblock \emph{arXiv preprint arXiv:2501.12948}, 2025.

\bibitem[Bertsch et~al.(2024)Bertsch, Ivgi, Alon, Berant, Gormley, and Neubig]{bertsch2024context}
Amanda Bertsch, Maor Ivgi, Uri Alon, Jonathan Berant, Matthew~R Gormley, and Graham Neubig.
\newblock In-context learning with long-context models: An in-depth exploration.
\newblock \emph{arXiv preprint arXiv:2405.00200}, 2024.

\bibitem[Khattab et~al.(2023)Khattab, Singhvi, Maheshwari, Zhang, Santhanam, Vardhamanan, Haq, Sharma, Joshi, Moazam, et~al.]{khattab2023dspy}
Omar Khattab, Arnav Singhvi, Paridhi Maheshwari, Zhiyuan Zhang, Keshav Santhanam, Sri Vardhamanan, Saiful Haq, Ashutosh Sharma, Thomas~T Joshi, Hanna Moazam, et~al.
\newblock Dspy: Compiling declarative language model calls into self-improving pipelines.
\newblock \emph{arXiv preprint arXiv:2310.03714}, 2023.

\bibitem[Opsahl-Ong et~al.(2024)Opsahl-Ong, Ryan, Purtell, Broman, Potts, Zaharia, and Khattab]{opsahl2024optimizing}
Krista Opsahl-Ong, Michael~J Ryan, Josh Purtell, David Broman, Christopher Potts, Matei Zaharia, and Omar Khattab.
\newblock Optimizing instructions and demonstrations for multi-stage language model programs.
\newblock \emph{arXiv preprint arXiv:2406.11695}, 2024.

\bibitem[Brown et~al.(2024)Brown, Juravsky, Ehrlich, Clark, Le, R{\'e}, and Mirhoseini]{brown2024large}
Bradley Brown, Jordan Juravsky, Ryan Ehrlich, Ronald Clark, Quoc~V Le, Christopher R{\'e}, and Azalia Mirhoseini.
\newblock Large language monkeys: Scaling inference compute with repeated sampling.
\newblock \emph{arXiv preprint arXiv:2407.21787}, 2024.

\bibitem[Wang et~al.(2024{\natexlab{a}})Wang, Cassano, Wu, Bai, Song, Nath, Han, Hendryx, Yue, and Zhang]{wang2024planning}
Evan Wang, Federico Cassano, Catherine Wu, Yunfeng Bai, Will Song, Vaskar Nath, Ziwen Han, Sean Hendryx, Summer Yue, and Hugh Zhang.
\newblock Planning in natural language improves llm search for code generation.
\newblock \emph{arXiv preprint arXiv:2409.03733}, 2024{\natexlab{a}}.

\bibitem[Wang et~al.(2024{\natexlab{b}})Wang, Yang, Li, Lu, Xu, Lin, Lin, Huang, and Wang]{wang2024scaling}
Xiyao Wang, Zhengyuan Yang, Linjie Li, Hongjin Lu, Yuancheng Xu, Chung-Ching Lin, Kevin Lin, Furong Huang, and Lijuan Wang.
\newblock Scaling inference-time search with vision value model for improved visual comprehension.
\newblock \emph{arXiv preprint arXiv:2412.03704}, 2024{\natexlab{b}}.

\bibitem[Shinn et~al.(2023)Shinn, Cassano, Gopinath, Narasimhan, and Yao]{shinn2023reflexion}
Noah Shinn, Federico Cassano, Ashwin Gopinath, Karthik Narasimhan, and Shunyu Yao.
\newblock Reflexion: Language agents with verbal reinforcement learning.
\newblock \emph{Advances in Neural Information Processing Systems}, 36:\penalty0 8634--8652, 2023.

\bibitem[Wang et~al.(2023)Wang, Xie, Jiang, Mandlekar, Xiao, Zhu, Fan, and Anandkumar]{wang2023voyager}
Guanzhi Wang, Yuqi Xie, Yunfan Jiang, Ajay Mandlekar, Chaowei Xiao, Yuke Zhu, Linxi Fan, and Anima Anandkumar.
\newblock Voyager: An open-ended embodied agent with large language models.
\newblock \emph{arXiv preprint arXiv:2305.16291}, 2023.

\bibitem[Saad-Falcon et~al.(2024)Saad-Falcon, Lafuente, Natarajan, Maru, Todorov, Guha, Buchanan, Chen, Guha, R{\'e}, et~al.]{saad2024archon}
Jon Saad-Falcon, Adrian~Gamarra Lafuente, Shlok Natarajan, Nahum Maru, Hristo Todorov, Etash Guha, E~Kelly Buchanan, Mayee Chen, Neel Guha, Christopher R{\'e}, et~al.
\newblock Archon: An architecture search framework for inference-time techniques.
\newblock \emph{arXiv preprint arXiv:2409.15254}, 2024.

\bibitem[Hu et~al.(2024)Hu, Lu, and Clune]{hu2024automated}
Shengran Hu, Cong Lu, and Jeff Clune.
\newblock Automated design of agentic systems.
\newblock \emph{arXiv preprint arXiv:2408.08435}, 2024.

\bibitem[Zhang et~al.(2024)Zhang, Xiang, Yu, Teng, Chen, Chen, Zhuge, Cheng, Hong, Wang, et~al.]{zhang2024aflow}
Jiayi Zhang, Jinyu Xiang, Zhaoyang Yu, Fengwei Teng, Xionghui Chen, Jiaqi Chen, Mingchen Zhuge, Xin Cheng, Sirui Hong, Jinlin Wang, et~al.
\newblock Aflow: Automating agentic workflow generation.
\newblock \emph{arXiv preprint arXiv:2410.10762}, 2024.

\bibitem[Peters and Schaal(2007)]{peters2007reinforcement}
Jan Peters and Stefan Schaal.
\newblock Reinforcement learning by reward-weighted regression for operational space control.
\newblock In \emph{Proceedings of the 24th international conference on Machine learning}, pages 745--750, 2007.

\bibitem[Jaderberg et~al.(2017)Jaderberg, Dalibard, Osindero, Czarnecki, Donahue, Razavi, Vinyals, Green, Dunning, Simonyan, et~al.]{jaderberg2017population}
Max Jaderberg, Valentin Dalibard, Simon Osindero, Wojciech~M Czarnecki, Jeff Donahue, Ali Razavi, Oriol Vinyals, Tim Green, Iain Dunning, Karen Simonyan, et~al.
\newblock Population based training of neural networks.
\newblock \emph{arXiv preprint arXiv:1711.09846}, 2017.

\bibitem[Barto(2021)]{barto2021reinforcement}
Andrew~G Barto.
\newblock Reinforcement learning: An introduction. by richard’s sutton.
\newblock \emph{SIAM Rev}, 6\penalty0 (2):\penalty0 423, 2021.

\bibitem[Shridhar et~al.(2020)Shridhar, Yuan, C{\^o}t{\'e}, Bisk, Trischler, and Hausknecht]{shridhar2020alfworld}
Mohit Shridhar, Xingdi Yuan, Marc-Alexandre C{\^o}t{\'e}, Yonatan Bisk, Adam Trischler, and Matthew Hausknecht.
\newblock Alfworld: Aligning text and embodied environments for interactive learning.
\newblock \emph{arXiv preprint arXiv:2010.03768}, 2020.

\bibitem[Yang et~al.(2023)Yang, Prabhakar, Narasimhan, and Yao]{yang2023intercode}
John Yang, Akshara Prabhakar, Karthik Narasimhan, and Shunyu Yao.
\newblock Intercode: Standardizing and benchmarking interactive coding with execution feedback.
\newblock \emph{Advances in Neural Information Processing Systems}, 36:\penalty0 23826--23854, 2023.

\bibitem[Jiang et~al.(2020)Jiang, Luketina, Nardelli, Minervini, Torr, Whiteson, and Rockt{\"a}schel]{jiang2020wordcraft}
Minqi Jiang, Jelena Luketina, Nantas Nardelli, Pasquale Minervini, Philip~HS Torr, Shimon Whiteson, and Tim Rockt{\"a}schel.
\newblock Wordcraft: An environment for benchmarking commonsense agents.
\newblock \emph{arXiv preprint arXiv:2007.09185}, 2020.

\bibitem[Reimers and Gurevych(2019)]{reimers-2019-sentence-bert}
Nils Reimers and Iryna Gurevych.
\newblock Sentence-bert: Sentence embeddings using siamese bert-networks.
\newblock In \emph{Proceedings of the 2019 Conference on Empirical Methods in Natural Language Processing}. Association for Computational Linguistics, 11 2019.
\newblock URL \url{http://arxiv.org/abs/1908.10084}.

\bibitem[Douze et~al.(2024)Douze, Guzhva, Deng, Johnson, Szilvasy, Mazar{\'e}, Lomeli, Hosseini, and J{\'e}gou]{douze2024faiss}
Matthijs Douze, Alexandr Guzhva, Chengqi Deng, Jeff Johnson, Gergely Szilvasy, Pierre-Emmanuel Mazar{\'e}, Maria Lomeli, Lucas Hosseini, and Herv{\'e} J{\'e}gou.
\newblock The faiss library.
\newblock \emph{arXiv preprint arXiv:2401.08281}, 2024.

\bibitem[Yao et~al.(2022)Yao, Chen, Yang, and Narasimhan]{yao2022webshop}
Shunyu Yao, Howard Chen, John Yang, and Karthik Narasimhan.
\newblock Webshop: Towards scalable real-world web interaction with grounded language agents.
\newblock \emph{Advances in Neural Information Processing Systems}, 35:\penalty0 20744--20757, 2022.

\bibitem[Yang et~al.(2018)Yang, Qi, Zhang, Bengio, Cohen, Salakhutdinov, and Manning]{yang2018hotpotqa}
Zhilin Yang, Peng Qi, Saizheng Zhang, Yoshua Bengio, William~W Cohen, Ruslan Salakhutdinov, and Christopher~D Manning.
\newblock Hotpotqa: A dataset for diverse, explainable multi-hop question answering.
\newblock \emph{arXiv preprint arXiv:1809.09600}, 2018.

\end{thebibliography}

\appendix
\appendix

\section{Key Agent Details}
\label{app:agent_details}

In Sec.~\ref{sec:preliminaries}, we establish an agent design that enables it to learn in-context from its own self-collected experiences. Here, we elaborate on a few key design decisions in our agent design:

\begin{itemize}
    \item \textbf{Standardized prompts}: we use the same simple, task-agnostic prompt templates for all tasks, rather than writing new prompts per task. These prompts are in App.~\ref{app:implementation}. Alternate appraoches incorporate domain-specific information into their prompts--we discuss these approaches in Appendices~\ref{app:automanual} and ~\ref{app:intercode_handcrafted}.
    \item \textbf{Two-level retrieval}: We retrieve trajectories at both trajectory level (for planning) and state level (for reasoning and action selection), enabling the agent to leverage both strategic patterns and situation-specific techniques. Database $\mathcal{D}$ contains self-collected trajectories, and retrieval is performed at the trajectory level for the initial plan $p$, and in the state-level observation-reasoning-action loop for both $r_t$ and $a_t$. 
    \item \textbf{Multi-key retrieval}: All retrieval is performed by KNN, with similarity metric defined as the average of cosine similarities across the specified `key' variables. For instance, in Line 3 of Alg.~\ref{alg:react_stepwise}, we retrieve from $\mathcal{D}$ using two keys: goal $g$ and plan $p$. We return similar trajectories based off the average of the cosine similarities of goals and plans when comparing each trajectory to the current trajectory. When doing state-level retrieval (Lines 7 and 10), we additionally find the most similar states within the selected trajectories via state-level key $o_t$ or $r_t$, then return a window of states around the most similar state. This is similar to the retrieval scheme in~\citep{kagaya2024rap}. See detailed pseudocode for retrieval in Alg.~\ref{alg:retrieval_pseudocode}.
    \item \textbf{Thought-based retrieval}: For the first step of a trajectory, we retrieve using the trajectory-level keys ($g$,$p$) as well as the current observation $o_1$ (Alg.~\ref{alg:react_stepwise}, line 6)--but for all subsequent steps we use reasoning $r_t$ as a key instead of observation $o_t$ (Alg.~\ref{alg:react_stepwise}, line 9). This approach, inspired by~\citet{zhou2024trad}, enables generalization across trajectories with similar reasoning, and similarity across natural-language $r_t$ can be handled by generic embedding functions more easily than potentially bespoke observations $o_t$. By retrieving at every step, we aim to retrieve the most relevant trajectories for each decision. 
    \item \textbf{Generic embedding mechanism}: Since $g$, $p$, and $r_t$ are all natural-language strings, we employ standard embeddings (all-MiniLM-L6-v2~\citep{reimers-2019-sentence-bert}) that generalize across domains without task-specific engineering.
\end{itemize}

\begin{algorithm}[t]
\caption{Multi-key Retrieval}
\begin{algorithmic}[1]
\Procedure{MultiKeyRetrieval}{$\mathcal{D}$, traj\_keys, state\_key, query, $k$, window\_size}
    \State $\text{similarities} \gets []$
    \For{each trajectory $\tau$ in $\mathcal{D}$}
        \State $\text{sim} \gets 0$
        \For{each key in traj\_keys}
            \State $\text{sim} \gets \text{sim} + \text{CosineSimilarity}(\text{query[key]}, \tau[\text{key}])$
        \EndFor
        \State $\text{sim} \gets \text{sim} / |\text{traj\_keys}|$ \Comment{Average similarity across trajectory keys}
        \State $\text{similarities}.\text{append}(\text{sim}, \tau)$
    \EndFor
    \State $\text{similar\_trajectories} \gets \text{TopK}(\text{similarities}, k)$
    
    \If{state\_key is not None} \Comment{State-level retrieval with window}
        \State $\text{windowed\_results} \gets []$
        \For{each trajectory $\tau$ in $\text{similar\_trajectories}$}
            \State $\text{state\_similarities} \gets []$
            \For{each state $s$ in $\tau.\text{states}$}
                \State $\text{state\_sim} \gets \text{CosineSimilarity}(\text{query[state\_key]}, s[\text{state\_key}])$
                \State $\text{state\_similarities}.\text{append}(\text{state\_sim}, s, \text{index}(s))$
            \EndFor
            \State $\_, \text{most\_similar\_state}, \text{idx} \gets \text{Max}(\text{state\_similarities})$
            \State $\text{window\_start} \gets \max(0, \text{idx} - \lfloor\text{window\_size}/2\rfloor)$
            \State $\text{window\_end} \gets \min(|\tau.\text{states}|, \text{idx} + \lceil\text{window\_size}/2\rceil)$
            \State $\text{windowed\_results}.\text{append}(\tau.\text{states}[\text{window\_start}:\text{window\_end}])$
        \EndFor
        \State \Return $\text{windowed\_results}$
    \Else
        \State \Return $\text{similar\_trajectories}$
    \EndIf
\EndProcedure
\end{algorithmic}
\label{alg:retrieval_pseudocode}
\end{algorithm}

\section{Additional Implementation Details}
\label{app:implementation}

\subsection{Prompt Templates}
\label{app:prompts}

Across all benchmarks, we use standardized prompt templates for the core components of our retrieval-based ReAct agent. 
The same templates were used across all benchmarks with no task-specific modifications.
These templates are intentionally minimalist, focusing on providing the necessary context and retrieved examples while avoiding task-specific prompt engineering. 

The templates are included below. Across all templates, the in-context examples follow the format specified in the prompt itself (for plan, the in-context examples are of form ``goal,plan'', etc): 

Plan:
\begin{mintedbox}{python}
system_prompt: 'You are an expert at generating high-level plans of actions to achieve a goal.\n Here is your action space: {action_space}.\n Here are some examples of goal,plan from episodes that successfully achieved similar goals: {examples}'
user_prompt: 'goal: {goal}\n plan: '
\end{mintedbox}

Reason:
\begin{mintedbox}{python}
system_prompt: 'You are an expert at reasoning about the most appropriate action to take towards achieving a goal.\n Here is your action space: {action_space}.\n Here are some examples of goal,plan,observation,reasoning,action from episodes that successfully achieved similar goals: {examples}'
user_prompt: 'goal: {goal}\n plan: {plan}\n trajectory: {trajectory}\n reasoning: '
\end{mintedbox}

Act:
\begin{mintedbox}{python}
system_prompt: 'You are an agent in an environment. Given the current observation, you must select an action to take towards achieving the goal: {self.goal}.\n Here is your action space: {action_space}.\n Here are some examples of goal,plan,observation,reasoning,action from episodes that successfully achieved similar goals: {examples}'
user_prompt: 'goal: {goal}\n plan: {plan}\n trajectory: {trajectory}\n action: '
\end{mintedbox}

\subsection{Retrieval Implementation}
\label{app:retrieval}

For all retrieval steps, we implement hybrid search across all the desired retrieval keys--ex. goal, plan, observation, reasoning. We return the top-$k$ examples by averaged distance across each of the keys. We implement a sliding window approach for state-level retrieval to enhance contextual relevance--we include the surrounding context (preceding and following states) up to a window of 5 steps to provide coherent episode fragments.

The retrieval mechanism is implemented using FAISS~\citep{douze2024faiss} for efficient similarity search as the database grows. We use exact nearest neighbor search. 

\subsection{Population-Based Training Details}
\label{app:pbt}

Our database-level curation approach maintains a population of 5 database instances. Each instance is initialized with the same set of human-provided exemplars. The population undergoes curation every time the database size doubles, and performance is evaluated on the tasks attempted since the previous doubling.

The replacement strategy follows standard population-based training practices: the bottom 20\% of databases (based on validation performance) are replaced with copies of the top 20\%. 

\subsection{Quality Metric Computation}
\label{app:quality_metric}

For exemplar-level curation, we track the retrieval patterns of each trajectory throughout the training process. For each task, we record:
1. Which trajectories were retrieved
2. How many times each trajectory was retrieved during the solution process
3. Whether the task was successfully completed

After completing all training tasks, we compute the quality metric $Q(\tau)$ for each trajectory $\tau$ as:

\begin{equation}
Q(\tau) = \frac{\sum_{i \in \mathcal{R}(\tau)} o_i \cdot f_i(\tau)}{\sum_{i \in \mathcal{R}(\tau)} f_i(\tau)}
\end{equation}

where $\mathcal{R}(\tau)$ is the set of tasks for which trajectory $\tau$ was retrieved, $o_i \in \{0, 1\}$ is the outcome of task $i$, and $f_i(\tau)$ is the retrieval frequency of trajectory $\tau$ during task $i$.

To ensure statistical significance, we only compute the quality metric for trajectories that were retrieved for at least 3 different tasks. For trajectories with insufficient retrieval data, we assign a neutral quality score equal to the average success rate across all tasks.

\subsection{Note on planning step}
Following the convention from RAP~\citep{kagaya2024rap}, we omit the planning step on benchmarks with short trajectory length (Intercode-SQL, Wordcraft). This planning step is valuable for maintaining long-horizon coherence on the ALFWorld benchmarks (30 steps), and is standard in prior ReAct-based agentic work~\citep{kagaya2024rap,zhao2024expel,fu2024autoguide}, whether the planning step is explicitly separate from reasoning, or incorporated into the first reasoning step. 

\section{Can Self-Collected Examples Improve a Fine-Tuned LLM Agent?}
\label{app:finetune}

We have shown that self-collected databases improve the performance of in-context LLM agents. Here, we test whether the same data can also benefit fine-tuning.

Using the OpenAI fine-tuning API, we fine-tune GPT-4o-mini on each benchmark using data from our best-performing database construction method: \trajbootstrap{}\dbexemplarcuration{}, collected over the full training set. We fine-tune using a simple ReAct-format prompt:

\begin{mintedbox}{python}
{
'system':'You are a ReAct agent that helps users accomplish tasks. Given a goal, you will receive observations about the environment and respond with your reasoning and actions. For each observation, first think through the problem step by step (Thought), then decide on an action (Action). Your actions should be clear, concise, and directly executable in the environment.',
'user':'Goal: {goal} \n Initial observation: {observations[0]}',
'assistant':'Thought: {reasoning[i]}\nAction: {action[i]}',
'user':'Observation: {observations[i+1]}',
...
}
\end{mintedbox}

We refer to the resulting fine-tuned model as ReAct-Finetune. To run the agent, we prompt it with a goal and initial observation, then alternate assistant messages (for reasoning and action) with user messages (for new observations).

Tab.~\ref{tab:finetune} shows that ReAct-Finetune slightly outperforms the in-context agent on ALFWorld (0.96 vs. 0.93) and Wordcraft (0.74 vs. 0.69), while performing slightly worse on InterCode-SQL (0.79 vs. 0.82). These results suggest that self-collected examples are effective not only for in-context prompting but also for creating competitive fine-tuned agents.

\begin{table}[t]
\centering
\begin{tabularx}{\textwidth}{l r r r}
\toprule
\textbf{Method} & \textbf{ALFWorld} & \textbf{InterCode-SQL} & \textbf{Wordcraft} \\
\midrule
\trajbootstrap{}\dbexemplarcuration{} & 0.93$\pm$0.03 & \textbf{0.82}$\pm$0.01 & 0.69$\pm$0.01 \\
ReAct-Finetune & \textbf{0.96}$\pm$0.01 & 0.79$\pm$0.01 & \textbf{0.74}$\pm$0.01 \\
\bottomrule
\end{tabularx}
\caption{\textbf{Trained on the same data, fine-tuned agent ReAct-Finetune is competitive with our best in-context approach.} This suggests that our self-collected data is effective not only for in-context prompting but also for creating fine-tuned agents. All values are averages over 5 trials.}
\label{tab:finetune}
\end{table}

\section{Can We Predict Agent Success Rates}
\label{app:prediction}

We have shown that increasing the number of self-collected examples improves agent performance. Here, we test whether the same examples can also predict performance on new tasks.

On InterCode-SQL and Wordcraft, task difficulty is partly observable from the goal $g$ and initial observation $o_1$. For InterCode-SQL, $g$ is a natural-language query. For Wordcraft, $g$ is the desired element and $o_1$ specifies available crafting elements. In contrast, ALFWorld task difficulty depends heavily on scene layout, which $g$ and $o_1$ do not reveal. We therefore exclude ALFWorld from this analysis.

We use the same embedding model as in retrieval (all-MiniLM-L6-v2~\citep{reimers-2019-sentence-bert}) to encode the concatenated string $[g; o_1]$. We train a calibrated Random Forest classifier to predict task success/failure, calibrating its outputs via 5-fold cross-validation with a learned sigmoid function. For each of 5 independent \trajbootstrap{} trials, we evaluate (1) the classifier’s AUROC on held-out tasks, and (2) its calibration.

\begin{figure}[t]
\centering
\begin{subfigure}[b]{0.32\textwidth}
\centering
\includegraphics[width=\textwidth]{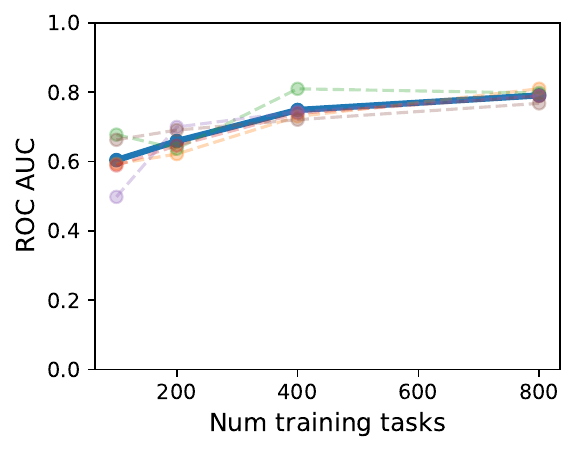}
\caption{InterCode-SQL}
\end{subfigure}
\begin{subfigure}[b]{0.32\textwidth}
\centering
\includegraphics[width=\textwidth]{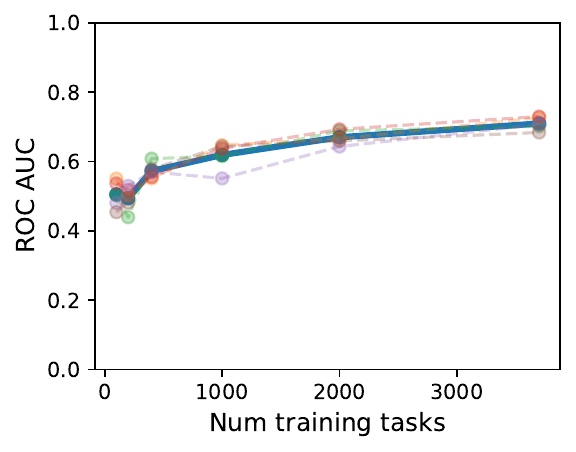}
\caption{Wordcraft}
\end{subfigure}
\caption{\textbf{AUROC of success prediction improves with more self-collected examples.} Performance continues to rise with increasing database size.}
\label{fig:auroc}
\end{figure}

As shown in Fig.~\ref{fig:auroc}, prediction performance improves as the database grows. For InterCode-SQL, AUROC rises from 0.60 (100 tasks) to 0.77 (800 tasks). Wordcraft shows a similar trend, improving from 0.50 (100 tasks) to 0.71 (4000 tasks). In both cases, predictive accuracy increases alongside task performance.

\begin{figure}[t]
\centering
\begin{subfigure}[b]{0.32\textwidth}
\centering
\includegraphics[width=\textwidth]{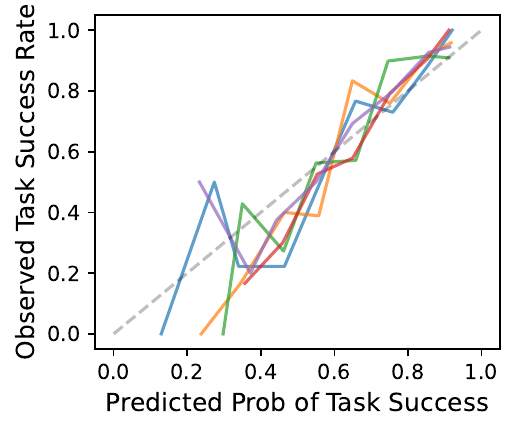}
\caption{InterCode-SQL}
\end{subfigure}
\begin{subfigure}[b]{0.32\textwidth}
\centering
\includegraphics[width=\textwidth]{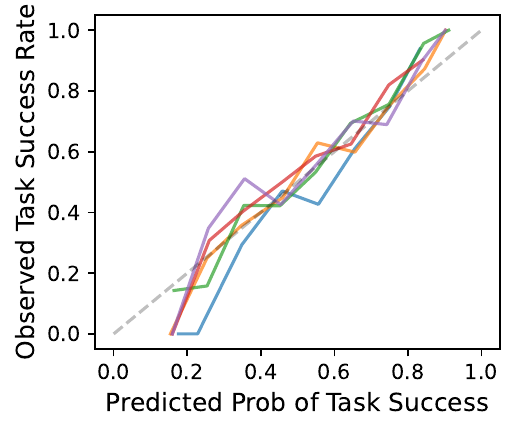}
\caption{Wordcraft}
\end{subfigure}
\caption{\textbf{Predicted probabilities are well-calibrated.} For both benchmarks, predicted and empirical success rates generally align.}
\label{fig:calibration}
\end{figure}

Fig.~\ref{fig:calibration} shows the calibration of the final classifiers (trained on all available training tasks). Predicted success probabilities closely match observed success rates, indicating well-calibrated models.

\section{Is it Possible to Bootstrap an Agent Without Initial Hand-Crafted Examples?}
\label{app:no_human}

Providing a small number of hand-crafted in-context examples is standard practice in the LLM agent literature~\citep{yao2023react,zhao2024expel,fu2024autoguide,chen2024automanual,kagaya2024rap}. 
However, what if we initialized \trajbootstrap{} with an empty database? 
In order to understand the value of the initial human-provided examples, we test Traj-Boostrap with and without the initial human-provided examples on Wordcraft. We refer to the variant initialized with an empty database as -Human-Examples. \trajbootstrap{}, initialized by default with a database of 5 human-provided trajectories for Wordcraft, achieves better performance with these starting examples than when initialized from an empty database (-Human-Examples). Performance still scales with database size for -Human-Examples--but in this case fails to reach the performance achieved via 5 human-provided examples, even after self-collecting trajectories on 4000 training tasks. On at least this one task, the initial human-provided trajectories shaped the reasoning and action patterns of the agent in a way that boosted the continual database construction process. We leave exploration of hand-crafting these in-context examples to future work. 

\begin{figure}[t]
    \centering
    \begin{subfigure}[b]{0.32\textwidth}
        \centering
        \includegraphics[width=\textwidth]{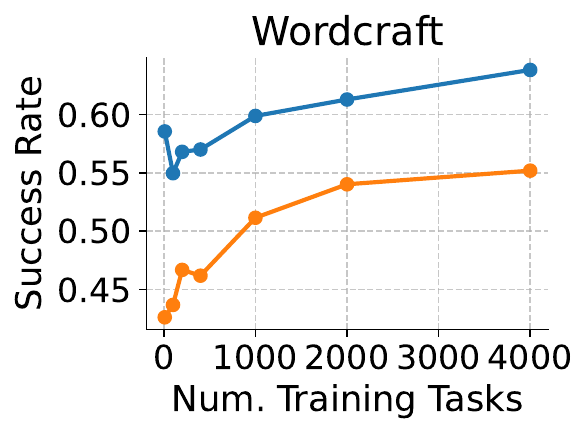}
        \includegraphics[width=0.75\textwidth]{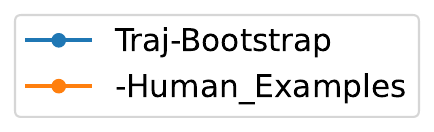}
        %\caption{Wordcraft}
        \label{fig:wordcraft_zero}
    \end{subfigure}
    \caption{\textbf{Ablating the value of initial human-provided examples, Wordcraft.} \trajbootstrap{}, initialized by default with a database of 5 human-provided trajectories for Wordcraft, achieves better performance with these starting examples than when initialized from an empty database (-Human-Examples). Performance still scales with database size for -Human-Examples--but in this case fails to reach the performance achieved via 5 human-provided examples, even after self-collecting trajectories on 4000 training tasks.}
    \label{fig:wordcraft_zero_results}
\end{figure}

\section{Key Details of Prior Agentic Approaches}

\subsection{How does Automanual Leverage Hand-Crafted Information}
\label{app:automanual}

Rather than learning from self-collected examples, an alternate approach to agent construction is to leverage practitioner domain knowledge.
Beyond implementing both a hierarchical learning system and code-based action spaces, Automanual~\citep{chen2024automanual} incorporates domain knowledge about the ALFWorld task into multiple components of the algorithm. In this section we include some code from the official Automanual GitHub to illustrate.

\paragraph{Observation spaces}: Automanual uses a modified observation space that enhances the ALFWorld string by adding two critical pieces of information: 1) The current location of the agent, 2) What the agent is currently holding. Both of these pieces of information typically have to be deduced from the trajectory of previous observations and actions, but Automanual tracks them explicitly:

\begin{mintedbox}{python}
if "Nothing happens" not in observation:
    self.last_obs = observation
    if "go to" in script:
        self.cur_loc = re.search(r'go to (\S+)', script).group(1)
        self.cur_loc_info = observation
    if "open" in script or "close" in script:
        self.cur_loc_info = observation
    if "take" in script:
        self.holding = re.search(r"(?<=take\s)(.*?)(?=\sfrom)", script).group(1)
        self.cur_loc_info = ""
    if "put" in script:
        self.holding = "nothing"
        self.cur_loc_info = ""
elif "go to" in script:
    loc = re.search(r'go to (\S+)', script).group(1)
    if loc == self.cur_loc: 
        observation = self.cur_loc_info
observation += f" You are at {self.cur_loc} and holding {self.holding}."
\end{mintedbox}

\paragraph{Action spaces}: in ALFWorld, any task typically involves three main components: 1) Searching for an object, 2) Performing an action with the object (heating, cooling, cleaning, etc.), 3) Placing the object somewhere. Automanual significantly simplifies both the search and placement operations by providing multi-action helper functions within its code-based action space:

\begin{mintedbox}{python}
# Define a helper method to find object that is needed
def find_object(agent, recep_to_check, object_name):
    for receptacle in recep_to_check:
        observation = agent.go_to(receptacle)
        # Check if we need to open the receptacle. If we do, open it.
        if 'closed' in observation:
            observation = agent.open(receptacle)
        # Check if the object is in/on the receptacle.
        if object_name in observation:
            object_ids = get_object_with_id(observation, object_name)
            return object_ids, receptacle
    return None, None

# Define a helper method to put object in/on the target receptacle
def go_to_put_object(agent, target_receptacle, object_id):
    observation = agent.go_to(target_receptacle)
    # check if target_receptacle is closed. If so, open it.
    if 'closed' in observation:
        observation = agent.open(target_receptacle)
    observation = agent.put_in_or_on(object_id, target_receptacle)
    return observation
\end{mintedbox}

\subsection{A note on training and test sets}
\label{app:train_test}

The distinction between how different techniques leverage data is crucial in understanding the generalization capabilities of LLM agents. We can categorize existing approaches based on how they treat training and test data:

\paragraph{Single-Task Optimization}: Some approaches focus exclusively on improving performance on a single task instance without concern for generalization. For example,~\citet{shinn2023reflexion} leverages feedback from failed attempts to incrementally improve performance on the same task. Similarly, search methods~\citep{brown2024large,wang2024scaling} expand the solution space for a specific problem instance. While these approaches can solve individual tasks, they don't transfer knowledge across different problems, essentially `overfitting` to a single instance.

\paragraph{Mixed Train-Test Evaluation}: Some recent work blurs training and test boundaries. For instance, RAP~\citep{kagaya2024rap} makes multiple passes over the same dataset, allowing the system to `learn` from some test examples before evaluating on others within the same set. This approach does not assess true generalization capability, as the model has indirect exposure to the test distribution during its learning phase. 

\paragraph{Full Train-Test Separation}: Several papers maintain a clear separation between training and test data: 1) ExpeL~\citep{zhao2024expel} extracts general rules from a training set of trajectories and applies them to entirely separate test tasks, 2) AutoGuide~\citep{fu2024autoguide} generates contextual guidelines from training experiences that are evaluated on distinct test scenarios. 3) AutoManual~\citep{chen2024automanual} constructs hierarchical `manuals` from training interactions that are then applied to novel test tasks.

Our approach similarly ensures that trajectories used for database construction come exclusively from designated training tasks, with evaluation conducted on a separate set of test tasks never seen during the database construction phase. This separation is essential for validating that the knowledge captured by the agent generalizes to new problems rather than memorizing specific solutions.

\section{Comparison to Hand-Crafted InterCode-SQL Agent}
\label{app:intercode_handcrafted}

\begin{table}[t]
    \centering
    \begin{tabularx}{0.9\textwidth}{l r r}
    \toprule
    \textbf{Method} & \textbf{Intercode-SQL Success Rate} \\
    \midrule
    GameSQL & 0.73 \\
    GameSQL+Cheat & \textbf{0.84} \\
    \midrule
    Fixed-DB & 0.74 \\
    \midrule
    \trajbootstrap{} & 0.79 \\
    \dbcuration{} & 0.78 \\
    \exemplarcuration{} & 0.81 \\
    \dbexemplarcuration{} & \textbf{0.82} \\
    \bottomrule
    \end{tabularx}
    \caption{ 
    \textbf{Comparison of agent success rates on InterCode-SQL: contextualizing the performance of Traj-Boostrap.} 
    Without cheats, the hand-crafted GameSQL agent (0.73) performs comparably to Fixed-DB (0.74). With handicap access to the database schema, GameSQL+Cheat (0.84) slightly outperforms \dbexemplarcuration{} (0.82). The boost from our database-construction techniques nearly matches the boost from providing the GameSQL agent with access to privileged database schema information. 
    }
    \label{tab:intercode_results}
\end{table}

We further contextualize the performance of \trajbootstrap{} by comparing to two hand-crafted agents on InterCode-SQL.
The Intercode-SQL paper~\citep{yang2023intercode} provides a hand-crafted agent, GameSQL to solve the task, and optionally provides the agent with a `handicap'--giving the agent information on all relevant parts of the database schema. We denote the assisted version as GameSQL+Cheat. Neither agent provides in-context examples, and both share a bespoke, hand-crafted prompt (see App.~\ref{app:gamesql}).

As seen in Tab.~\ref{tab:intercode_results}, Fixed-DB performs similarly to GameSQL (0.74 vs 0.73), and the performance of our best method, \dbexemplarcuration{}, approaches the performance of GameSQL+Cheat (0.82 vs 0.84). Therefore, our database-construction techniques lift the performance of a generic ReAct-style agent nearly as much as the lift provided to the hand-crafted agent via `handicap' access to the database schema.

\section{Benchmark Details}
\label{app:benchmarks}

\subsection{ALFWorld}
\label{app:alfworld}

ALFWorld \citep{shridhar2020alfworld} is a text-based environment that aligns with embodied tasks, allowing agents to navigate and manipulate objects through textual commands. We use the standard ALFWorld benchmark consisting of ~3500 training tasks and 134 out-of-distribution test tasks across 6 task categories:
\begin{itemize}
    \item Pick \& Place: Find and move objects to specified locations
    \item Clean \& Place: Find, clean, and place objects
    \item Heat \& Place: Find, heat, and place objects
    \item Cool \& Place: Find, cool, and place objects
    \item Pick Two \& Place: Find and move two objects to a specified location
    \item Look at Object: Find an object and examine it under light
\end{itemize}

Following \citep{kagaya2024rap}, for the ALFWorld benchmark we perform similarity search over task categories in addition to the other retrieval keys (goal, plan, observation, action). We do this to follow the convention in this prior work. 

For our initial human-provided exemplars, we used the 18 successful trajectories (3 per task category) provided by~\citet{zhao2024expel}. These trajectories were used to initialize all database instances. 

The success criteria for ALFWorld tasks are defined by the environment and require the agent to satisfy all conditions specified in the goal. For example, in a `Heat \& Place` task, the agent must find the target object, place it in the microwave, turn on the microwave, and finally place the heated object at the specified destination. Both Autoguide and Automanual allow 50 actions for task completion--but choosing to employ "reasoning" counts as an action. Since we force our agent to reason at every step, we allow our agents (Fixed-DB, \trajbootstrap{} and variants) only 30 steps for task completion (on Autoguide and Automanual, the agent does not reason in practice at most steps ex. in a search procedure). 

For this benchmark, we do not provide an action space string to the LLM, relying purely on the in-context examples to communicate the action space. 

\subsection{InterCode-SQL}
\label{app:intercode}

InterCode-SQL \citep{yang2023intercode} is an interactive coding environment for evaluating language agents' SQL programming abilities. We use a subset of the InterCode benchmark focusing on SQL query generation, built upon the Spider SQL dataset. Of the 1034 tasks in the dataset, we randomly assign 800 tasks to train and the remaining 234 tasks to test.

Each task provides a natural language query request. The agent must generate a syntactically correct SQL query that retrieves the requested information. The agent must first execute queries to understand the database schema. The environment provides feedback on syntax errors and execution results, but the agent is only allowed to submit a solution once. 

The success criteria for InterCode-SQL tasks require the agent to submit a solution query within 10 steps. The environment executes the query and compares the results against a ground-truth reference.

For our initial human-provided exemplars, we collected 10 human-created trajectories for 10 randomly-selected training tasks. These trajectories were used to initialize all database instances. For all solved trajectories, we append the solution query to the goal string--since the goal of the task is to `discover' this query through interacting with the SQL database. 

We used the following action space string for InterCode-SQL:
\begin{lstlisting}[style=promptstyle]
Your action space is outputting valid mysql commands to solve the sql task.
You will be evaluated on the Latest Standard Output.
If you believe the latest observation is the final answer, you can complete the task by running 'submit' by itself.
You have 10 iterations to solve the task.
Follow the syntax and logical flow from the provided examples exactly.
\end{lstlisting}

\subsection{Wordcraft}
\label{app:wordcraft}

Wordcraft~\citep{jiang2020wordcraft} is a simplified adaptation of the game Little Alchemy, where agents must combine elements to create new elements through multi-step processes. We randomly select 4000 training tasks and 500 test tasks from the subset of tasks requiring up to 2 steps to solve, with the train-test split separating the tasks into disjoint sets of goal elements. 

The agent starts with a set of elements and must discover combinations that creates a particular target element specified in the goal. The environment provides feedback on successful combinations and updates the available elements accordingly.

The success criteria for Wordcraft tasks require the agent to create the target element within 4 steps, while the minimum solution length is up to 2 steps.

For our initial human-provided exemplars, we collected 4 human-annotated trajectories from randomly-selecting training tasks. These trajectories were used to initialize all database instances. We collected fewer initial trajectories for Wordcraft than for InterCode-SQL (4 vs 10) since Wordcraft is a slightly simpler task, requiring up to 4 steps for task completion while InterCode-SQL requires up to 10.

We used the following action space string for Wordcraft:
\begin{lstlisting}[style=promptstyle]
Output strings with the names of the two entities we would like to combine in this step.
\end{lstlisting}

\subsection{Note on Benchmark Selection}

We selected three sequential decision-making benchmarks that cover different reasoning challenges—\textbf{ALFWorld} \citep{shridhar2020alfworld} tests text-based navigation and object manipulation, \textbf{InterCode-SQL} \citep{yang2023intercode} tests interactive code generation, and \textbf{Wordcraft} \citep{jiang2020workcraft} tests compositional reasoning.

While prior works~\citep{zhao2024expel,fu2024autoguide} test on WebShop~\citep{yao2022webshop}, we encountered bugs in generating achievable goals on the full benchmark (confirmed by https://github.com/princeton-nlp/WebShop/issues/43) and identified tasks with incorrect rewards.

We excluded QA benchmarks (HotPotQA~\cite{yang2018hotpotqa}, etc.) because performance depends on information retriever quality and LLM self-evaluation efficacy, two factors that would confound our study of LLM Self-Improvement. We plan to test our algorithms on QA benchmarks in future work.

\section{Computational Resources}
\label{app:compute}

All experiments were conducted using the following computational resources:

\begin{itemize}
    \item 1 NVIDIA A5000 GPU (24GB memory) for embedding computation
    \item 64GB RAM
\end{itemize}

The majority of computation was spent on OpenAI API calls for the LLM-based decision-making. Database operations including embedding computation, storage, and retrieval accounted for less than 5\% of the total computation time.

For embedding computations, we used all-MiniLM-L6-v2~\citep{reimers-2019-sentence-bert}.

For LLM inference, we used the OpenAI API for GPT-4o-mini, which required approximately:
\begin{itemize}
    \item 2,000,000 API calls for ALFWorld
    \item 200,000 API calls for InterCode-SQL
    \item 500,000 API calls for Wordcraft
\end{itemize}

The total cost of API usage was approximately \$3,000 USD.

\section{GameSQL Prompt}
\label{app:gamesql}

\citet{yang2023intercode} write this hand-crafted prompt for the GameSQL agent:

\begin{mintedbox}{python}
'''
{self.language}Env` is a multi-turn game that tests your ability to write
a {self.language} command that produces an output corresponding to a natural language query.

## GAME DESCRIPTION
At the start of this game, you are given a natural language query describing some
desired output (i.e. "Find the first name of a student who have both cat and dog pets").
Aside from the natural language query, you have no information about the tables you have access to.

The game will be played in a series of turns. Each turn, you can submit a {self.language} command.
You will then get a response detailing the output of your {self.language} query along with a reward
that tells you how close your {self.language} command is to the correct answer.

The goal of this game is to write a {self.language} command that gets a reward of 1. The game will automatically
terminate once you get a reward of 1.

## INPUT DESCRIPTION
Each turn, you can submit a {self.language} command. Your {self.language} command should be formatted as follows:

```{self.language}
Your {self.language} code here
```

Your {self.language} command can help you do one of two things:
1. Learn more about the tables you have access to
2. Execute {self.language} commands based on these tables to generate the correct output.

## OUTPUT DESCRIPTION
Given your {self.language} command input, `{self.language}Env` will then give back output formatted as follows:

Output: <string>
Reward: <decimal value between 0 and 1>

The output is a string displaying the result from executing your {self.language} query.
The reward is a decimal value between 0 and 1.

## REWARD DESCRIPTION
The reward should be interpreted as a ratio. It tells you how many rows your {self.language}
command outputted correctly compared to the correct answer.

## RULES
1. Do NOT ask questions. Your commands are fed directly into a SQL compiler. 

## STRATEGY
You are free to play as many turns of the game as you'd like to inspect tables
and develop your {self.language} command.

The best strategy for this game is to first write {self.language} commands that help you learn
about the tables that you have access to. For instance, in a SQL environment, you might use `SHOW TABLES`
and `DESC <table name>` to learn more about the tables you have access to.

Once you have a good understanding of the tables, you should then write {self.language} commands
that would answer the natural language query using the tables you have access to.
'''
\end{mintedbox}

\end{document}